\documentclass[journal]{IEEEtran}
\usepackage{amsmath,amsfonts}
\usepackage{algorithmic}
\usepackage{algorithm}
\usepackage{array}
\usepackage[caption=false,font=normalsize,labelfont=sf,textfont=sf]{subfig}
\usepackage{textcomp}
\usepackage{stfloats}
\usepackage{url}
\usepackage{verbatim}
\usepackage{graphicx}
\usepackage{cite}
\usepackage{mathrsfs}
\usepackage{amssymb}
\usepackage{footnote}
\usepackage{multirow}
\usepackage{tabularx}
\usepackage{booktabs}
\usepackage{utfsym}
\usepackage{bm}
\usepackage{mathrsfs}
\begin{document}

\title{LOIS: Looking Out of Instance Semantics\\ for Visual Question Answering}

\author{Siyu~Zhang,
        Yeming~Chen,
        Yaoru~Sun,
        Fang~Wang,
        Haibo~Shi,
        Haoran~Wang
\thanks{Siyu Zhang, Yeming Chen, Yaoru Sun, and Haoran Wang are with the Department of Computer Science and Technology, Tongji University, Shanghai 201804, China (e-mail:zsyzsy@tongji.edu.cn;2130769@tongji.edu.cn; yaoru@tongji.edu.cn; wanghaoran\_tj@tongji.edu.cn)}
\thanks{Fang Wang is with Department of Computer Science, Brunel University, Uxbridge UB8 3PH, UK.}
\thanks{Haibo Shi is with School of Statistics and Management, Shanghai University of Finance and Economics, Shanghai 200433, China.}
\thanks{(\textit{Corresponding author: Yaoru Sun.})}
}

\markboth{}%
{Shell \MakeLowercase{\textit{et al.}}: A Sample Article Using IEEEtran.cls for IEEE Journals}


\maketitle

\begin{abstract}
Visual question answering (VQA) has been intensively studied as a multimodal task that requires effort in bridging vision and language to infer answers correctly. Recent attempts have developed various attention-based modules for solving VQA tasks. However, the performance of model inference is largely bottlenecked by visual processing for semantics understanding. Most existing detection methods rely on bounding boxes, remaining a serious challenge for VQA models to understand the causal nexus of object semantics in images and correctly infer contextual information. To this end, we propose a finer model framework without bounding boxes in this work, termed \emph{Looking Out of Instance Semantics (LOIS)} to tackle this important issue. LOIS enables more fine-grained feature descriptions to produce visual facts. Furthermore, to overcome the label ambiguity caused by instance masks, two types of relation attention modules: 1) intra-modality and 2) inter-modality, are devised to infer the correct answers from the different multi-view features. Specifically, we implement a mutual relation attention module to model sophisticated and deeper visual semantic relations between instance objects and background information. In addition, our proposed attention model can further analyze salient image regions by focusing on important word-related questions. Experimental results on four benchmark VQA datasets prove that our proposed method has favorable performance in improving visual reasoning capability. 
\end{abstract}


\section{Introduction}
\IEEEPARstart{V}{isual} question answering (VQA) \cite{b1}, is an analytical reasoning task that connects two different modalities: computer vision (CV) and natural language processing (NLP)\cite{b2}, \cite{b3}. It is challenging as it requires correctly inferring text questions related to the contents of the images. Text questions mostly contain many sub-questions in various visual aspects, such as questions about object detection ``\emph{How many books are there?}'' and scene classification ``\emph{Is it cloudy?}''. In addition, there are also many complex questions about spatial relations reasoning ``\emph{What is the shape of the object closest to the large cylinder?}'' and commonsense reasoning ``\emph{Why is the boy crying?}''. These problems require rich semantic understanding from vision and language systems. Hence, devising a robust VQA system to deal with cross-modality interaction relations has proved to be crucial. 

Recent VQA methods focused on cross-modality fusion \cite{b4}, which exploited the CNN/RNN structure and combined the vision-language features.  With further development, attention-based methods \cite{b5}, \cite{b6}, \cite{b7} were used to find key clues for the correct answers. Most of the models have been greatly improved and have achieved high accuracy. In spite of this, these models still suffer from a semantic gap in visual understanding. As such, some critical challenges still need to be solved. On the one hand, the current models struggle to balance the effects of fusion-based multimodal mechanisms. On the other hand, the answers rely more on the relations to the text questions instead of reasoning upon image contents, which leads the model to generate spurious biases \cite{b8} in answer distributions. For instance, to answer the text question about ``\emph{What color are these bananas?}'', most answers are more likely to be predicted as ``\emph{yellow}'' than ``\emph{green}''. It explicitly reveals the inadequacy of learning image contents and semantic associations, so that inference from language modules easily overrides image semantics to generate fake understanding answers \cite{b9}.

\begin{figure}[!t]
\centering
\includegraphics[width=3.45in]{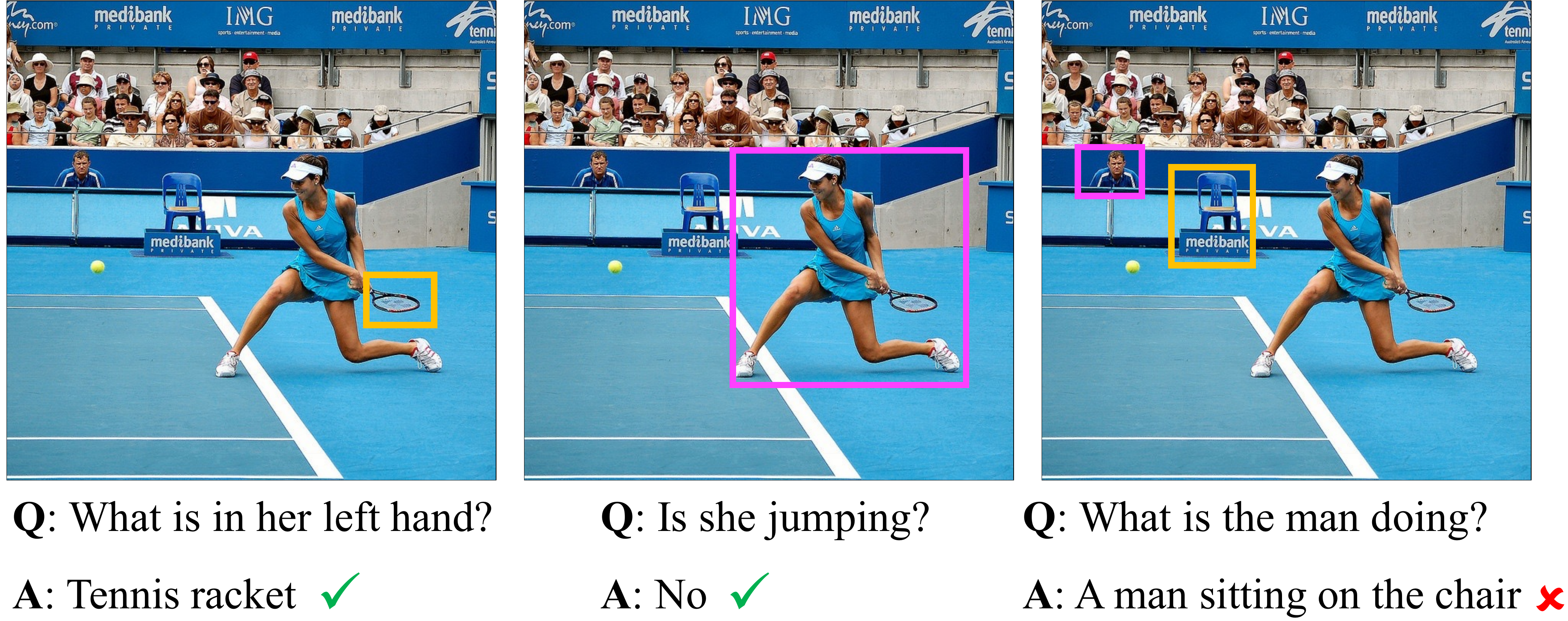}
\caption{Some examples for inferring object relations with different types. However, for an observer, the model may not be able to provide the correct answers due to ignoring the background semantic information of the objects.}
\label{fig_1}
\end{figure}

\IEEEpubidadjcol
For visual feature representation, existing methods mainly rely on off-the-shelf object detectors to deliver a set of bounding boxes or region-based features. Bounding boxes have been widely optimized and processed for their simplicity in representing object locations. For instance, the work of Anderson \textit{et al.} \cite{b10} utilized a combined bottom-up and top-down attention mechanism. In particular, the class of R-CNNs \cite{b11}, as an object detection method, was also applied to establish a closer link between vision and language tasks. Object detection methods based on bounding boxes have become the mainstream method in most VQA techniques. However, it is difficult for these methods to deeply understand both the image contents and semantic context information. Firstly, since the edge features of the foreground objects are obscured by background features, many unrelated feature regions are concentrated. As a result, it fails to accurately distinguish specific instance semantic features. Furthermore, human vision can effortlessly locate and perceive objects by their flexible attention boundaries rather than rough bounding boxes. Secondly, visual understanding of images is limited to the predefined categories for regions, and object contextual semantics mentioned by text questions are ignored. As shown in Fig.~\ref{fig_1}, the model obtained a wrong answer as ``\emph{a man sitting on the chair}'' unrelated to the contextual semantics. Thirdly, low spatial resolution features make it difficult to handle the details at object boundaries, leading to incorrect object predictions. More recently, some studies have revisited the grid visual feature representations for VQA. For instance, Jiang \textit{et al.} \cite{b12} attempted to learn grid features as an alternative to the widely used bounding box features. The immediate benefit was inference speed, which skipped all of the computationally expensive region-based steps in the VQA pipeline. However, it might also focus on irrelevant context or partial objects, and the performance was below the requirements. Huang \textit{et al.} \cite{b13} developed Pixel-BERT, an end-to-end visual language embedding framework for aligning text and pixel-level semantic connections. It turns out that most examples more focus on supporting concepts, and fail in providing higher accuracy (\emph{especially on counting questions}). Even though the existing  vision-language-based methods have achieved great progress, the image understanding module still needs to further mine the deep semantic information.

Unlike previously thought, the goal of our work is to improve the vision module for expressive feature representation. To this end, we design a multimodal relation attention model based on the ground-truth instance semantic segmentation, since its inherent advantages are as follows: (i) instance objects are optimized for finer edge features in a simple and uniform manner; (ii) the presented relation attention module can enhance long-range representation (i.e., \emph{contextual semantic}) in feature maps to alleviate redundancy of information between the local instance and the global background regions; (iii) multimodal semantic features with intra-modality and inter-modality are fully reconciled to infer the correct answers. Due to a lack of understanding of objects and relations within a larger context in VQA, few efforts have been devoted to analyzing and tackling the semantic association-based instance segmentation strategy. Hence, it needs good guidance for academic study to move forward meaningfully. Specifically, we propose \emph{looking out of instance semantics (LOIS)}, a finer, natural, rigorous, and bounding box-free model framework. Note that the limitation of the label ambiguity \cite{b14} is a serious challenge for extending VQA. Our solution yields two efficient relation attention modules. One is developed for learning multi-view correlations, and the other integrates the understanding of image contents and text questions to obtain answers by using inter-modality attention. To validate the effectiveness of the proposed framework, we offer more details on implementation and carry out extensive experiments on the VQA-v1 \cite{b1}, VQA-v2 \cite{b15}, COCO-QA \cite{b16}, and VQA-CP v2 (VQA under Changing Priors) \cite{b17} datasets. To the best of our knowledge, we are the first to apply instance semantic detection in the VQA tasks. 

The main contributions can be concluded as follows:

\begin{itemize}[\IEEEsetlabelwidth{Z}]
\item We introduce a novel framework, termed LOIS, to model instance masks in the VQA, which can describe more fine-grained edge features. Within our approach, there is no need for bounding box detection.
\item Our framework learns effective multi-view visual attributes, which leverage precise semantic features and sophisticated spatial relations. In addition, we make a trade-off between the local instance and the global background features.
\item To explore the high-level interactions between multi-view feature correlations and language domains, we further reconcile two efficient relation attention modules, intra-modality, and inter-modality, to learn the complementary knowledge and reinforce reasoning for the correct answers.
\item Our work highlights the vision bottleneck (i.e., \emph{visual facts}) to alleviate the influence of language priors on model decisions. More importantly, the proposed model is evaluated on four challenging benchmark datasets and achieves promising performance.\\
\end{itemize}

The remainder of this work is organized as follows. Section~II introduces related work of VQA. Our designed LOIS is mainly described in detail in Section~III, and attention mechanisms are presented, which contain the relation reasoning between the vision and questions. In Section~IV, the experimental results and further analysis are provided. Finally, Section~V summarizes this paper. 
\section{Related Work}
\subsection{Feature Extraction} 
Visual question answering has emerged as a typical multimodal task. Unlike other visual captioning \cite{b18}, \cite{b19} and cross-modal information retrieval (CMIR) \cite{b20}, \cite{b21} tasks, VQA requires comprehensive analysis and reasoning over both the image contents and natural language questions. Thus, VQA plays a crucial role in many applications, such as autonomous driving, blind people interaction \cite{b22}, medical assistance \cite{b23}, etc. A straightforward VQA strategy was to fuse visual and text questions for a joint embedding. Malinowski \textit{et al.} \cite{b24} proposed Neural-Image-QA, an end-to-end architecture that trained jointly based on the CNNs and LSTMs to generate correct answers. Initially, concatenation, element-wise multiplication, or addition operations were the most popular fusion strategies. Later on, more complex fusion methods \cite{b25} were used to improve joint vector representation. Nevertheless, these VQA strategies have shown limited performance. 

Recently, detection-based VQA methods become a mainstream task. Even so, visual semantics detection remains a challenge on the visual side, ranging from detecting pixel-wise semantics to inferring abstract scene semantics for the whole image. To distinguish foreground objects from background contexts, object detection predicts the corresponding semantic concepts by localizing a bounding box around each instance. Many well-known detectors, such as YOLO, RetinaNet, R-CNNs, and DETR Transformers-based, have been proposed to address these challenges. For instance, Teney \textit{et al.} \cite{b26} presented a joint embedding VQA framework, which generated a multi-label classifier over a set of candidate answers by employing Faster R-CNN pre-trained on the Visual Genome dataset to focus on specific elements. To reduce the noise and learn the interdependency between pairwise feature elements, Wu \textit{et al.} \cite{b27} proposed plug-and-play differential networks, where image features were encoded using Faster R-CNN, and GRU was used for question feature extraction. In \cite{b28}, Zhang \textit{et al.} introduced a knowledge-based augmentation network, which improved the fusion of richer visual and knowledge representations to obtain an accurate answer. With the development of object detection, instance segmentation is progressed to precisely segment each instance based on localizations. Xie \textit{et al.} \cite{b29} employed a new framework based on the polar representation, which converted instance segmentation to dense distance regression and instance center classification tasks. Chen \textit{et al.} \cite{b30} devised a flexible blender module for instance-level dense prediction tasks, and their blender can be added to most object detectors with modest computational overhead. In contrast to these common binary classification strategies in instance segmentation, we explore LOIS applying instance semantic representation to downstream tasks of VQA.

\subsection{Attention-based Models} 
For a given image, dozens of objects need to be extracted. Furthermore, both the spatial and semantic features of these objects are involved in the questions. It becomes important to correctly analyze the question, and quickly find the image contents related to the corresponding question. In light of this, most of the devised VQA methods succeed depending on attention mechanisms.

Existing visual attention methods \cite{b31}, \cite{b32} have been proposed to extract salient features from key regions of an image. Concretely, they can be classified into detection-based and free-form-based strategies. For the detection-based strategy, Wu \textit{et al.} \cite{b33} designed object-difference attention for VQA tasks that compared objects explicitly by a difference operator for calculating the attention distribution. To describe the more sophisticated trinary relations, Peng \textit{et al.}\cite{b34} introduced a composed relation attention network. In the work of Guo \textit{et al.} \cite{b6}, a re-attention model  was provided to help visual attention learning by minimizing consistency loss. For the methods based on free-form attention mechanisms, Yang \textit{et al.} \cite{b35} developed a stacked attention network to gradually filter out noise, in which the answer was progressively inferred by querying the image multiple times. In \cite{b36}, Li \textit{et al.} presented a relation-aware graph attention network, which attempted to explore multitype visual relations via a graph attention mechanism. By reasoning relations between image parts and questions, a recurrent attention unit was utilized in \cite{b37} to help the model process information sequentially. 

To strengthen reasoning capability and answer more complex questions, most existing attention-based methods have been developed as a co-attention mechanism to jointly analyze the key information in the questions and images. Peng \textit{et al.} \cite{b38} achieved a word-to-region attention network, which bridged the semantic gap between relevant regions and keywords. Besides, the experiment explicitly revealed that the keywords in the questions were important for locating visual regions. Liu \textit{et al.} \cite{b39} performed two effective adversarial learning of supervised attention modules. Specifically, both detection-based and free-form-based attention networks were utilized to learn multi-view features for answer inference. Most recently, several studies have discussed whether the above methods can improve performance. By employing a self-adaptive neural module transformer, Zhong \textit{et al.} \cite{b40} encoded question features to enhance multimodal fusion.  Combining the mutual attention module, Zheng \textit{et al.} \cite{b5} further proposed a novel method to achieve semantic correspondence between questions and images for remote sensing VQA. To improve performance and interpretability, a multimodal relation attention module was introduced in \cite{b41} to explore the complicated relations between visual and text questions. Different from previous VQA methods, we separately encode visuals and questions using two types of attention mechanisms. In particular, image features from different views are reinforced in visual attention modules to tackle the language priors.
\begin{figure}[!t]
\centering
\includegraphics[width=2.75in]{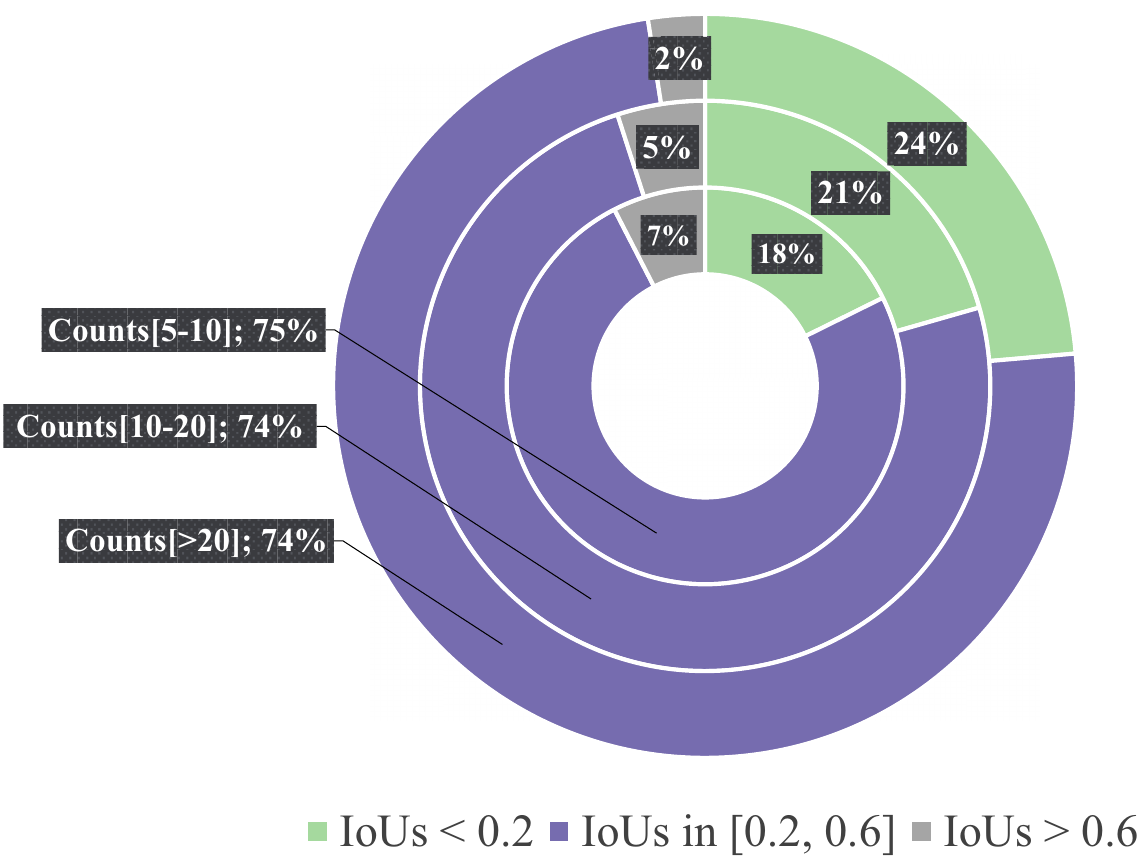}
\caption{IoUs statistical results for different instance object counts.}
\label{fig_2}
\end{figure}
\begin{figure*}[!t]
\centering
\includegraphics[width=7in]{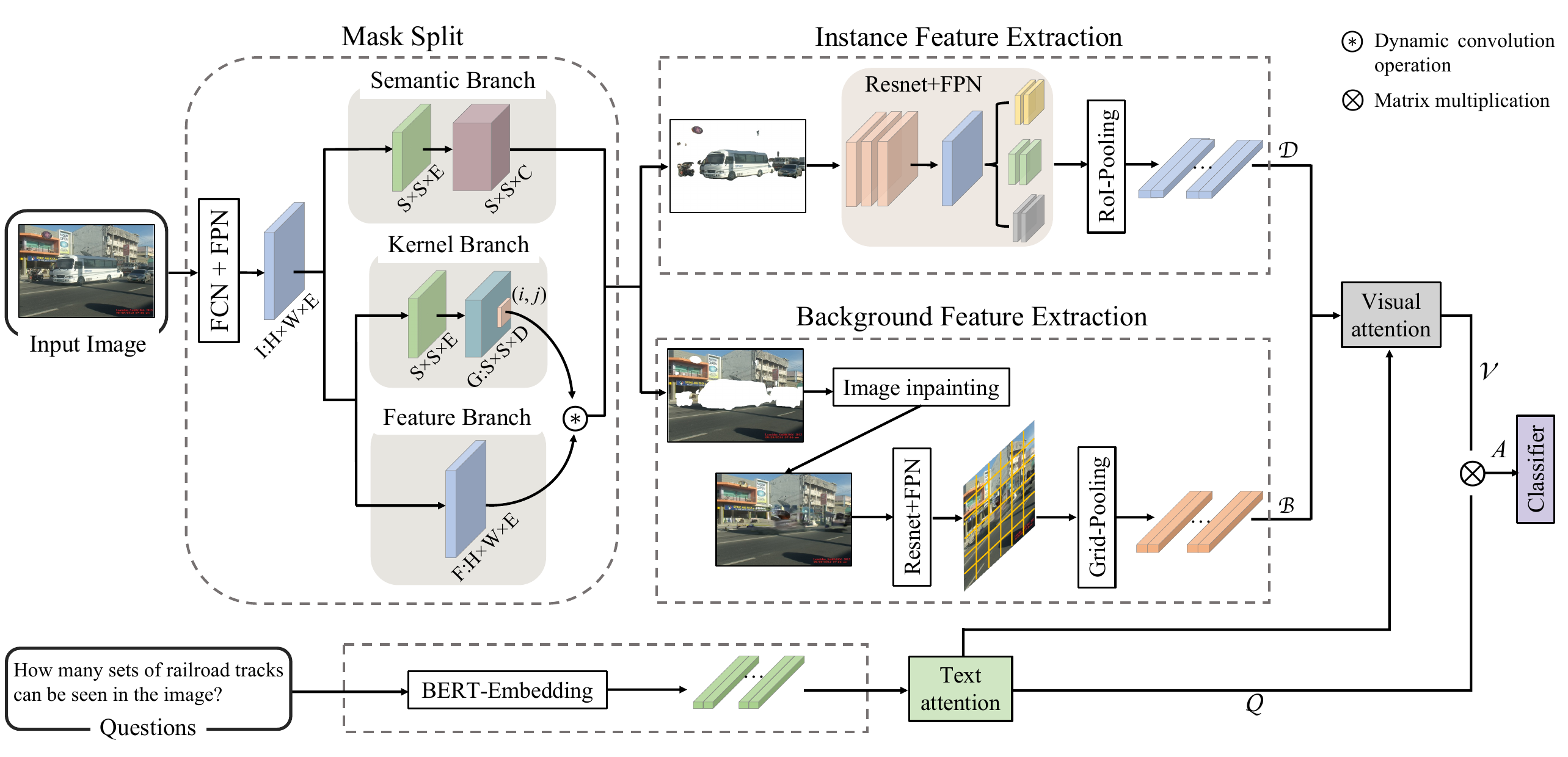}
\caption{Overview of the LOIS framework. (a) shows the instance mask learning process, which can be divided into three branches: semantic learning, convolution kernel learning, and feature learning. (b) is the instance feature extraction module, where the Resnet-101 is an underlying backbone to fuse FPN. In (c), we show the background feature extraction. Here, we perform an image inpainting network with GAN to prevent the shape of masks from generating clues.}
\label{fig_3}
\end{figure*}
\section{LOIS for visual question answering}
\subsection{Problem Statement}
Visual question answering problems involve describing information from the images. Yet, current methods are focused on  off-the-shelf visual detectors while less on visual facts and semantics. In particular, the bounding box-based methods have been widely employed in recent trends. However, as mentioned in Fig.~\ref{fig_2}, intersection over unions (IoUs) are computed to measure the proportion of edge overlaps between different numbers of instances (e.g., the number of objects ranges from [5$-$10], [10$-$20] and [$>$20]) in each sample. The statistical results indicate varying degrees of overlap in the extracted object edge information based on the bounding box when IoUs in [0.2, 0.6] or [$>$0.6]. Hence, it is easy to produce a negative impact on the visual facts. Instance segmentation is one of the fundamental object detection tasks, which aims at distinguishing different semantic categories. However, the major challenge is that visual detection or segmentation is insufficient to understand the holistic scene for accurate VQA. In contrast, the VQA system requires the provision of arbitrary questions related to content meanings in images, but it has received less attention in recent VQA works. Therefore, we propose a finer, pixel-level, and bounding box-free VQA system for instance semantics in this work to overcome this crucial problem. Moreover, the relations of instance objects with each other and the overall scene are considered to infer answers.

Intuitively, the overview of our LOIS framework is illustrated in Fig.~\ref{fig_3}. First, the image is input to learn instance mask. Then, we respectively obtain the foreground instance feature $\mathcal{D}$ and background forms $\mathcal{B}$. Next, the BERT \cite{b42} model is utilized to encode questions for extracting question features. After that, we introduce a reinforced attention mechanism to analyze the relations between multi-view images and questions. Finally, the inferred essential features from the visual image $\mathcal{V}$ and the text question $\mathcal{Q}$ are fed into a classifier for answer prediction.

\subsection{Feature Representation Component}
\textit{1) Instance semantics detection:} The proposed LOIS, as a pixel-level detection system, aims to extract image instance semantic features for VQA tasks, without embedding learning or bounding boxes. Instance regions are further detected on the basic concepts of SOLO v2 \cite{b43} in our work. In our framework, two simultaneous category-aware prediction tasks are mainly provided, instance detection and corresponding semantic probability prediction. Specifically, to consider the center location category, an image $\mathcal{V}$ is first conceptually divided into $S\times S$ grid cells (e.g., $S=12$), which causes the instance mask of $S^2=144$ channels to be generated at most. Note that the output channel $k^{th}$ is responsible for detecting instance masks at position ($i$, $j$), where $k=i\cdot S + j$. Then, the semantic category probabilities for each object instance are predicted through the $C$-dimensional output space $S\times S\times C$. The mask learning process contains mask features and a mask kernel. Next, the input pyramid feature $\mathcal{F}_{I}\in \mathbb{R}^{H_I\times W_{I}\times E}$ is aligned to the space $S\times S\times E$ (e.g., $E=64$), and the mask kernel $\mathcal{G}\in \mathbb{R}^{S\times S\times D}$ is obtained after performing the convolution operation of four convolution layers and $3\times3\times D$ convolution. As for $1\times1$ convolution with $E$ input channels, the kernel weight is set to $D=E$. When performing a $3\times3$ convolution, we will obtain $D=9E$. After that, instance mask prediction can be achieved through a dynamic convolution operation, which is presented as follows:
\begin{equation}
\setlength{\abovedisplayskip}{12pt}
\setlength{\belowdisplayskip}{12pt}
\label{eq:1}  
M_{i,j}= \mathcal{G}_{i,j}\circledast \mathcal{F}     
\end{equation}
where $\mathcal{F}\in \mathbb{R}^{H\times W\times E}$ refers to the mask feature, $\mathcal{G}_{i,j}\in \mathbb{R}^{1\times1\times E}$ represents the mask kernel of the $1\times1$ convolution at location ($i$, $j$), and $M_{i,j}\in \mathbb{R}^{H\times W}$ represents the final instance mask. $\circledast$ denotes the dynamic convolution operation.

During this process, there will be at most $S^2$ masks for each prediction level. Not all masks represent different instance objects. To distinguish between different instance objects that are represented by the same mask, only one mask with the highest confidence should be retained in the network output. To detect effective instance features from different semantic objects, a decay factor is employed by matrix non-maximum suppression (NMS) to reduce the confidence of redundant masks. The corresponding expression is provided as follows:

\begin{equation}
\label{eq:2}  
IoU_{a,b} = \frac{a\cap b}{a\cup b}  
\end{equation}
\begin{equation}
\label{eq:3}  
\Phi_a = \mathop{min}\limits_{\forall s_k>s_a} (1-IoU_{k,a})
\end{equation}
where $IoU_{a,b}$ is the intersection over union of masks $a$ and $b$. $s_k$ and $s_a$ are denoted as the confidences of mask $k$ and $a$, respectively. $\Phi_a$ represents the positive correlation function of mask $a$ with the $IoU$. $IoU_{k,a}$ indicates the decay function of mask $k$ and $a$.

For the above purpose, the prediction score is updated in a single step by the penalty factor, so that the best prediction mask is retained by setting the threshold. The penalty factor $penalty_{b}^{*}$ will obey the following rules:
\begin{equation}
\setlength{\abovedisplayskip}{12pt}
\setlength{\belowdisplayskip}{12pt}
\label{eq:4}  
\begin{split}
penalty_{b}^{*} &= \mathop{min}\limits_{\forall s_a>s_b}\frac{1-IoU_{a,b}}{\Phi_a } \\ 
s_b&\gets s_b \cdot penalty_{b}^{*}  
\end{split}
\end{equation}
where $s_b$ is the confidence of mask $b$.
\begin{figure}[!t]
\centering
\subfloat[]{\includegraphics[width=1.11in]{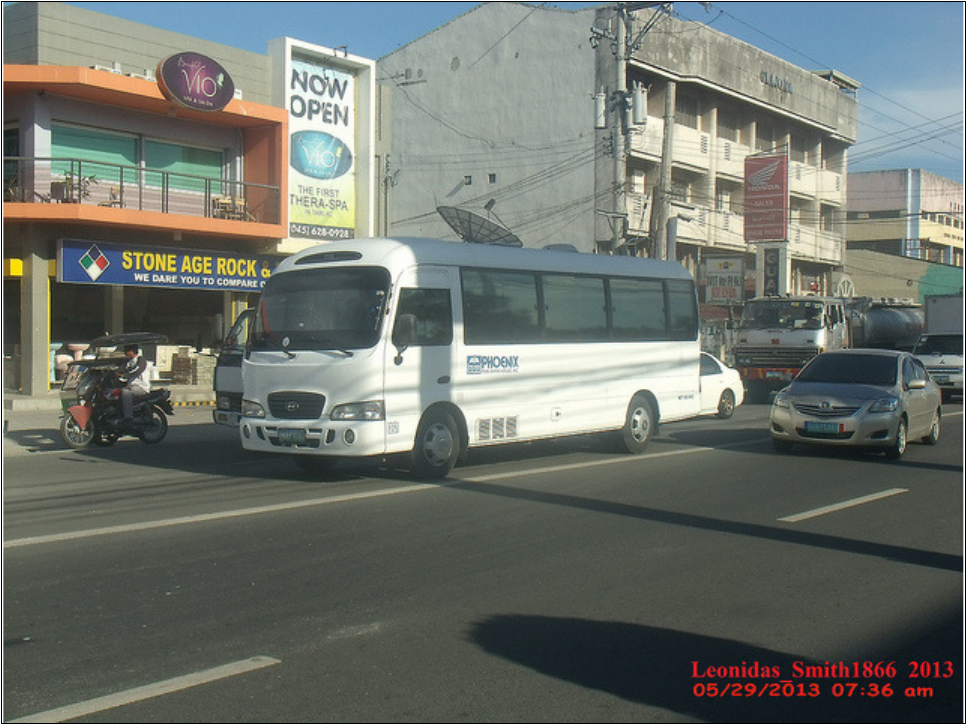}
\label{fig_a}}
\hfil
\subfloat[]{\includegraphics[width=1.11in]{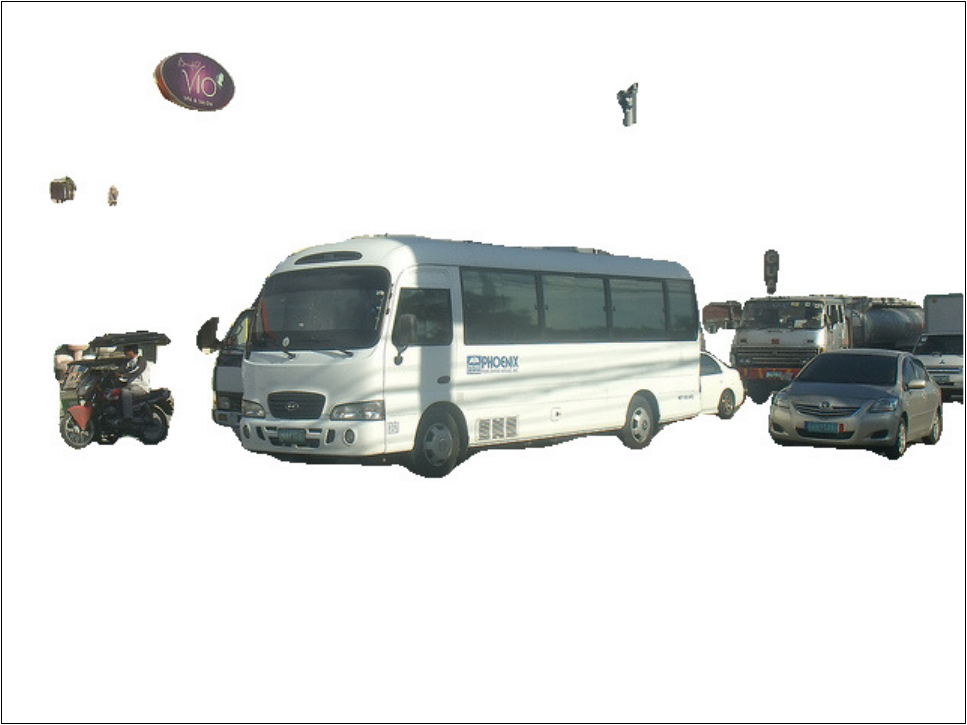}
\label{fig_b}}
\hfil
\subfloat[]{\includegraphics[width=1.11in]{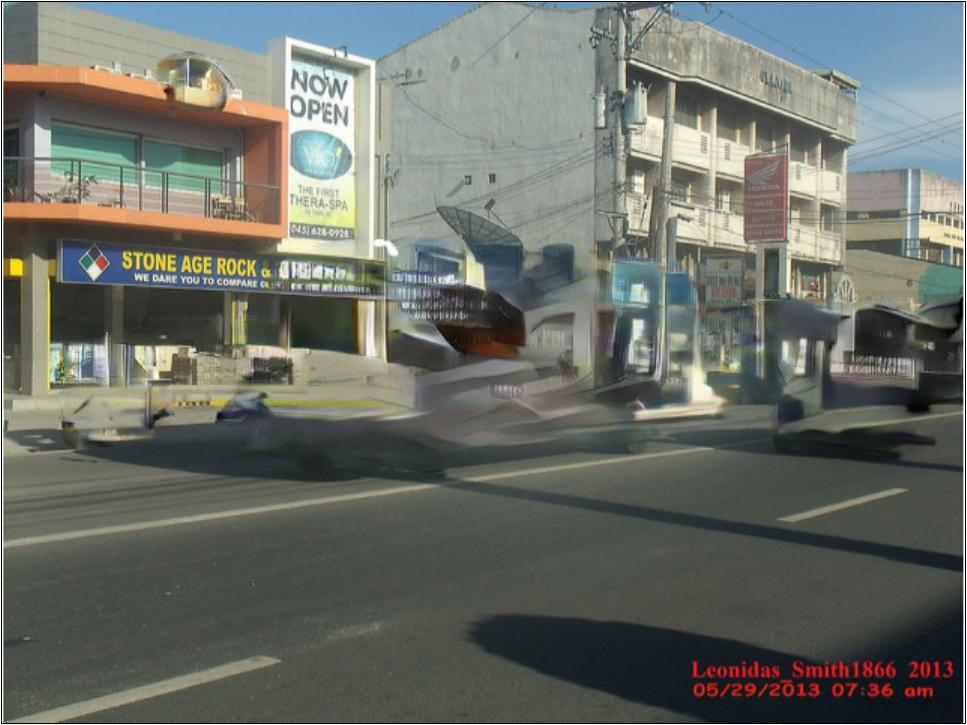}
\label{fig_c}}
\caption{Object instances and background features with different views: (a) Original image, (b) instance object selection, (c) background inpainting.} \label{fig_4}
\end{figure}

\textit{2) Image semantic features applied in VQA:} 
Differing from the spatial layout binary classification strategy adopted in the previous instance segmentation, our proposed model needs to learn deep semantic features from detected instance objects for the VQA tasks. To capture the effective edge distribution from different views, the detected instance masks are separated to obtain finer-grained features. The instance mask $\mathcal{D}$ with high confidence in one image is selected as follows:
\begin{equation}
\setlength{\abovedisplayskip}{12pt}
\setlength{\belowdisplayskip}{12pt}
\mathcal{D} = \uppercase\expandafter{\romannumeral2}(\sum\nolimits_{S_{i,j}>mask\_threshold}M_{i,j})
\label{eq:5} 
\end{equation}
where $M_{i,j}$ represents each separated instance mask location ($i$, $j$). Function $II$ represents the boolean values true or false. It is worth noting that different instance masks are fused into one feature map to reduce complex computation and unnecessary redundant information. Then, the corresponding background mask $\mathcal{B}$ is derived as follows:
\begin{equation}
\setlength{\abovedisplayskip}{12pt}
\setlength{\belowdisplayskip}{12pt}
\mathcal{B} = GAN(\mid \mathcal{D}-\mathcal{I}\mid\otimes \mathcal{V})
\label{eq:6} 
\end{equation}
where $\mathcal{V}$ and $\mathcal{I}$ denote the original image and the identity matrix, respectively. Then, $\otimes$ represents matrix multiplication. Meanwhile, to prevent the model from getting cues from the shape of the masks, a GAN-based inpainting network \cite{b44} is used to process these masks and generate realistic images.
\begin{figure}[!t]
\centering
\subfloat[]{\includegraphics[width=1.125in]{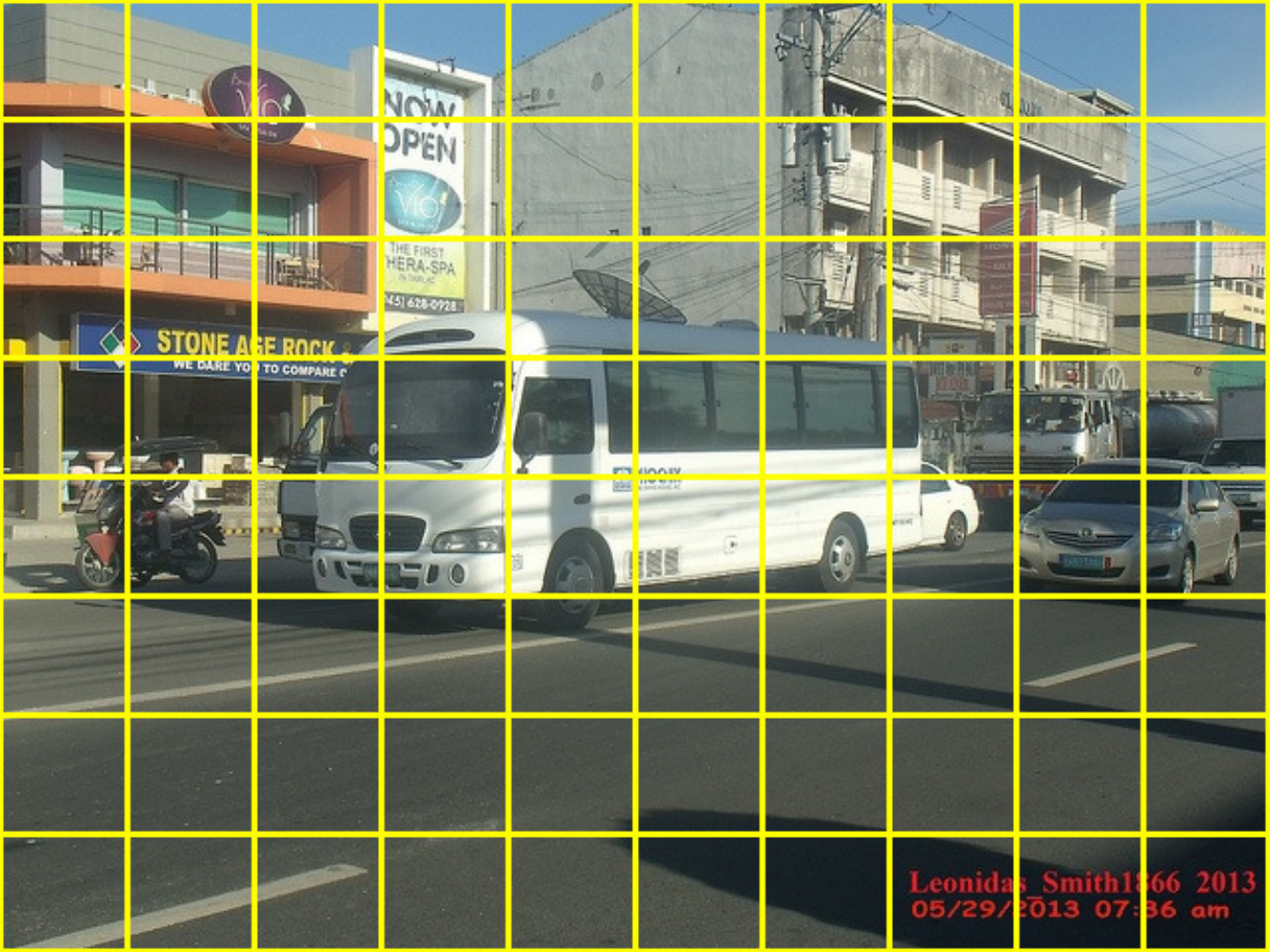}}
\hfil
\subfloat[]{\includegraphics[width=1.125in]{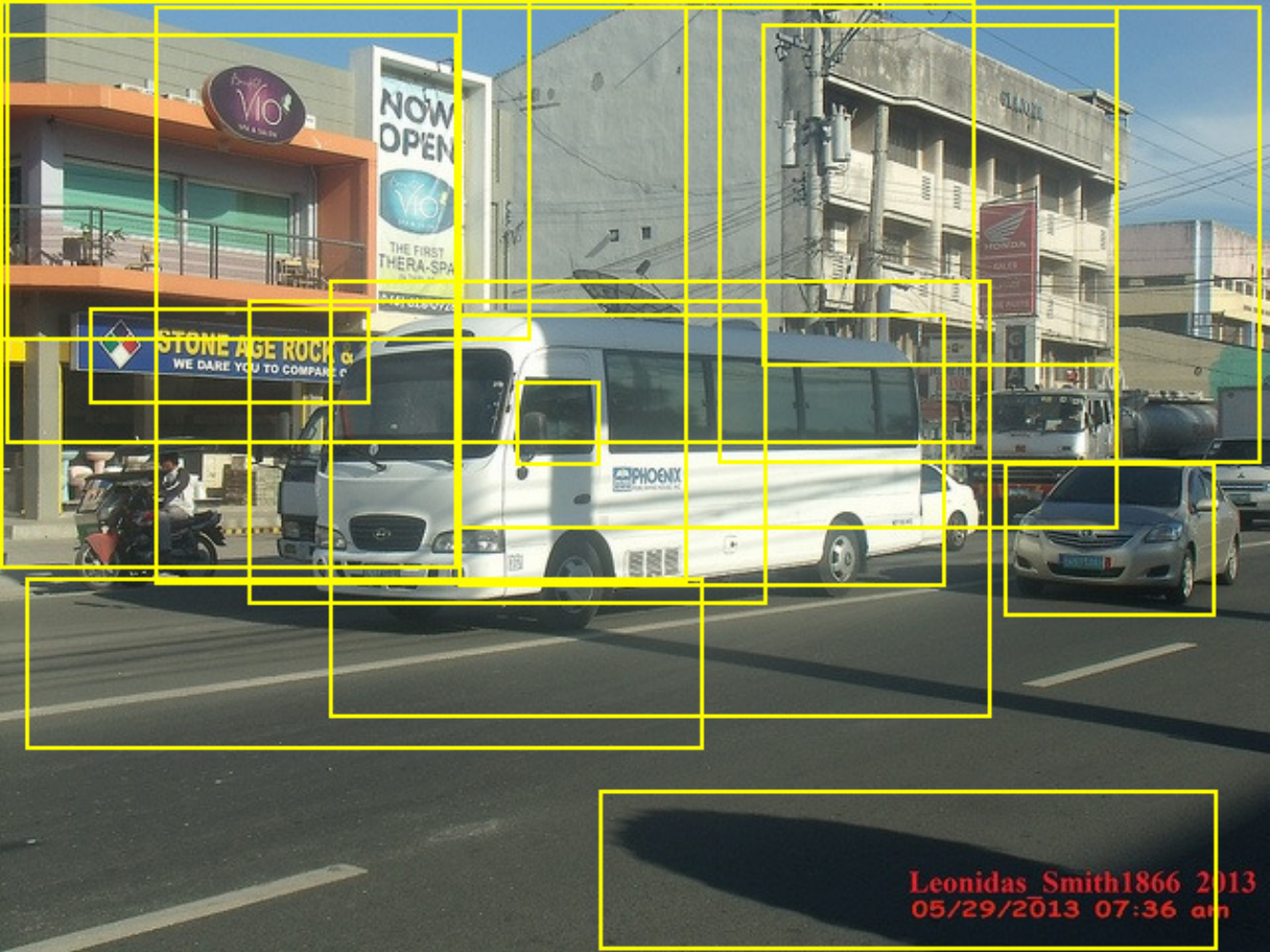}}
\hfil
\subfloat[]{\includegraphics[width=1.125in]{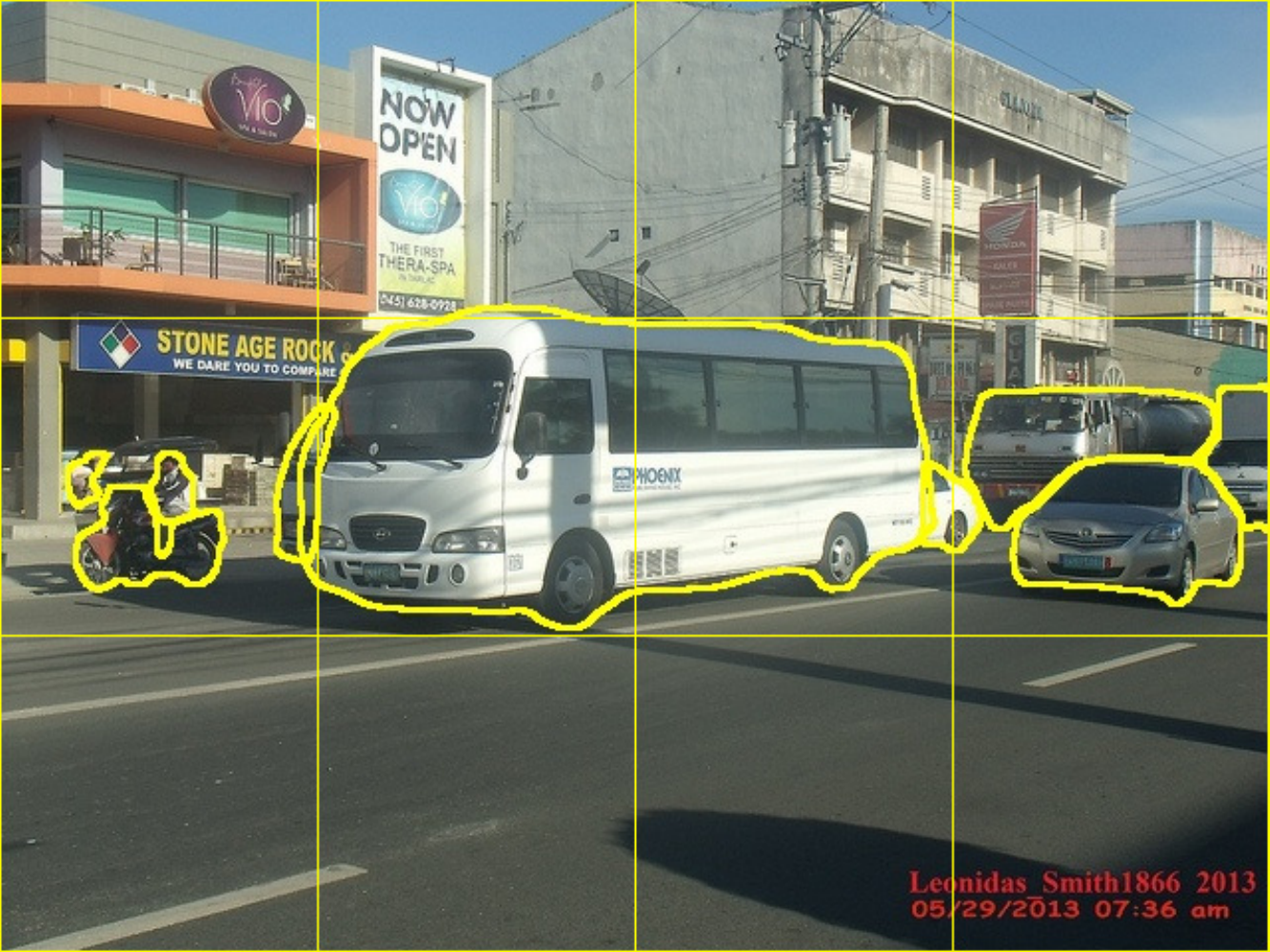}}
\caption{Uniform grid-based image regions of equal size (a), bounding box-based detection regions (b), instance-based semantic regions (c).} \label{fig_5}
\end{figure}
Since the specific objects in the background contexts are difficult to describe, it is inaccurate to predict based on the location information alone. We model the background contextual information by utilizing the uniform grid \cite{b12}, which fully takes into account the contextual semantic information other than the instance objects. In this way, the instance semantic features $\mathcal{D}\in \mathbb{R}^{L\times\rho}$ and the background features $\mathcal{B}\in \mathbb{R}^{L\times\phi}$ are obtained. Where $\rho$ and $\phi$ indicate the numbers of different features. $L$ represents the dimensionality of each feature vector. Here, we align these dimensions with instance features to show the effectiveness and simplicity of our instance modeling framework. Fig.~\ref{fig_4} provides different views of one image. Where the left (a) presents the original image. The middle (b) represents candidate object instances, which are identified from the visual images. Then, the right (c) shows the inpainted background feature maps. As shown, our proposed framework can provide high resolution for finer mask prediction to deal with the details at object boundaries. Notably, our designed detection process does not depend on bounding boxes compared with most of the detection methods. In other words, through the operation of the convolution matrix, the existence of regular boundaries on the retained location information is examined. But in fact, the valid values of the instance edge features are easily captured. Moreover, the feature values of regions outside the instance mask are set to 0, which does not affect the extraction of instance objects in this work. 

We select instance regions of an image instead of the bounding box-based regions and the uniform grid-based regions. The specific difference between them is shown in Fig.~\ref{fig_5}. Obviously, the proposed instance-based detection regions are more natural and representative than the others. We argue that instance semantic features are effective on benchmarks. For the reasoning abilities of the model in answering visual questions, a grid-based method for background features can lead to significant speed-ups and achieve stronger performance. We hope that the proposed framework can trade-off between the local instance regions and the global background from different multi-view features. In addition, it is interesting to note that the degree of interaction between the instance masks and the corresponding background form is a crucial factor influencing the VQA model. By analyzing from different independent perspectives, some slight differences may exist in the model. Hence, we put forward the relation attention module, which is discussed in detail in \textit{Sec.~C}. 

\subsection{Attention Module}
\begin{figure}[!t]
\centering
\includegraphics[width=3.42in]{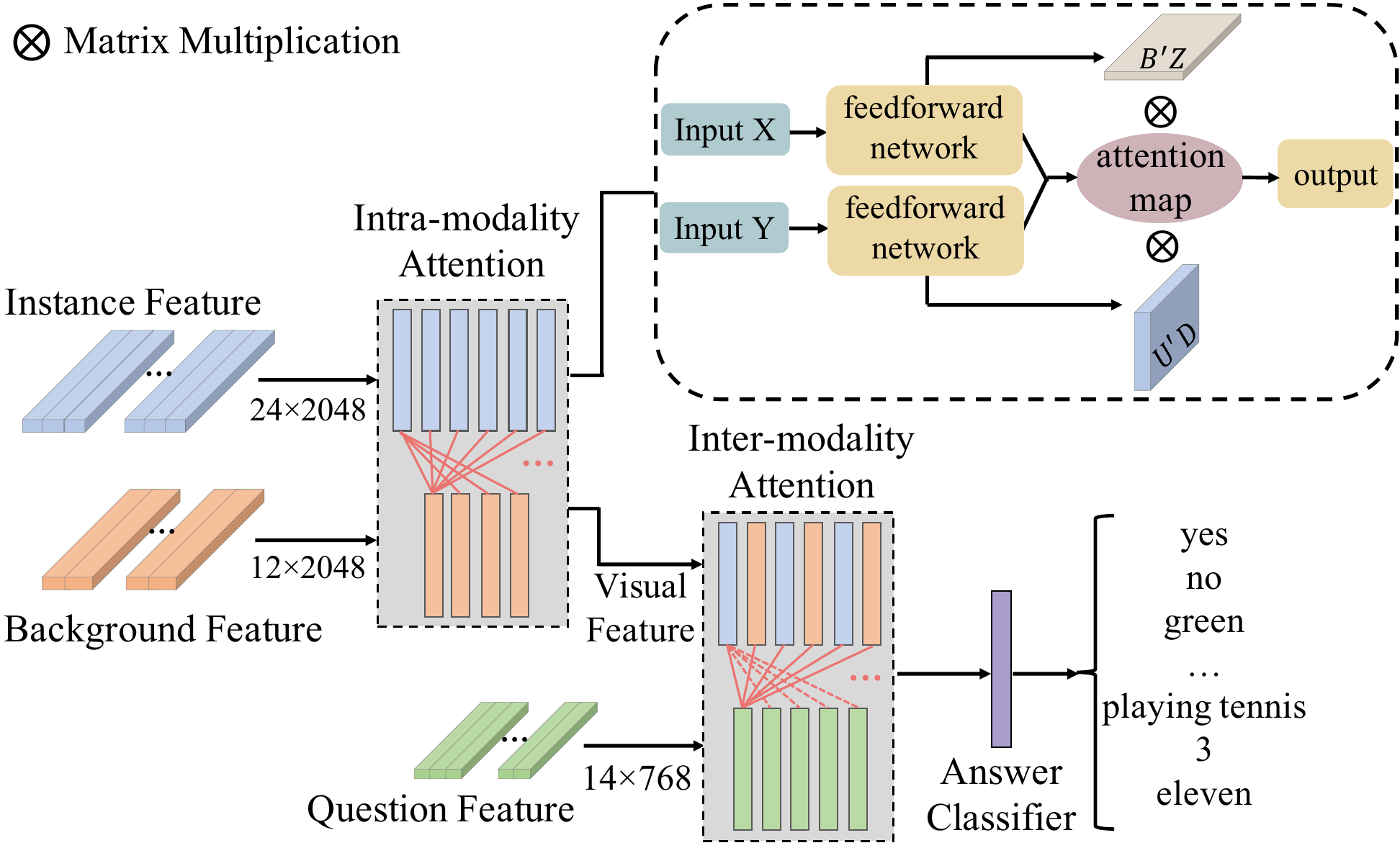}
\caption{Illustration of the proposed relation attention model with intra- and inter-modality.}
\label{fig_6}
\end{figure}

In VQA tasks, it is essential to provide a more holistic representation of images. Previous to this, several effective studies have suggested that two or more object relations are explored to enhance model reasoning capability. However, it is tough to apply question answering to real images due to its emphasis on reasoning. More critically, inherent object semantics are ignored. Different from previous model structures, we present two models of relation attention mechanisms to answer complicated scene questions, that is, intra-modality and inter-modality. The whole attention pipeline is illustrated in Fig.~\ref{fig_6}. Each component is elaborated upon below.

\textit{1) Intra-modality relation Attention:}
Considering to improve the accuracy of VQA, we argue that semantic features between multi-view should be learned. We assume two multi-channel inputs,\textit{i.e.}, instance detection features $\mathcal{D}\in \mathbb{R}^{L\times\rho}$ and background features $\mathcal{B}\in \mathbb{R}^{L\times\phi}$, consisting of $\rho=|\{\mathbf x_i\}|$ and $\phi=|\{\mathbf y_i\}|$vectors, respectively. Two groups of input channels $\mathcal{D}$ and $\mathcal{B}$, into the same dimension $L$, can be expressed as follows: 
\begin{equation}
\setlength{\abovedisplayskip}{12pt}
\setlength{\belowdisplayskip}{12pt}
\widetilde{\mathcal{D}} =\tanh(\mathcal{D}^{T}W_{1}^\prime+b_1), W_{1}^\prime\in \mathbb{R}^{L\times\rho}, b_{1}\in \mathbb{R}^{\rho}
\label{eq:7} 
\end{equation}
\begin{equation}
\widetilde{\mathcal{B}} =\tanh(\mathcal{B}^{T}W_{2}^\prime+b_2), W_{2}^\prime\in \mathbb{R}^{L\times\phi}, b_{2}\in \mathbb{R}^{\phi}
\label{eq:8} 
\end{equation}
where $W_1^\prime\in\mathbb{R}^{L\times\rho}$ and $W_2^\prime\in\mathbb{R}^{L\times\phi}$ are defined as the trainable weight matrices. $b_1\in \mathbb{R}^{\rho}$ and $b_2\in \mathbb{R}^\phi$ represent bias parameters. Then, we construct a bilinear attention map, which in our case is:
\begin{equation}
\setlength{\abovedisplayskip}{12pt}
\setlength{\belowdisplayskip}{12pt}
\mathcal{O}_{intra,r}^\prime=(\widetilde{\mathcal{D}}^{T}U^\prime)_{r}^{T}\alpha(\widetilde{\mathcal{B}}^{T}Z^\prime)_r
\label{eq:9} 
\end{equation}
where $\widetilde{\mathcal{D}}\in\mathbb{R}^{L\times\rho}$,$\widetilde{\mathcal{B}}\in\mathbb{R}^{L\times\phi}$, and $U^\prime\in\mathbb{R}^{L\times\gamma}$,$Z^\prime\in\mathbb{R}^{L\times\gamma}$. $r$ is used to represent the $r$-th column of matrices, and $\alpha\in\mathbb{R}^{\rho\times\phi}$ denotes a bilinear weight matrix. Be aware that at most $\gamma$-rank bilinear pooling exists for $\mathcal{O}_{intra}^\prime\in\mathbb{R}^{\gamma}$. Eq.~(\ref{eq:9}) can be rewritten as:
\begin{equation}
\setlength{\abovedisplayskip}{12pt}
\setlength{\belowdisplayskip}{12pt}
\label{eq:10}
\begin{split}
\mathcal{O}_{intra,r}^\prime&=\sum\nolimits_{i=1}^{\rho}\sum\nolimits_{j=1}^{\phi}\alpha_{i,j}(\widetilde{\mathcal{D}}_i^{T}U_r^{\prime})({Z^{\prime}}_r^{T}\widetilde{\mathcal{B}}_j) \\
                            &=\sum\nolimits_{i=1}^{\rho}\sum\nolimits_{j=1}^{\phi}\alpha_{i,j}\widetilde{\mathcal{D}}_i^{T}(U_r^\prime {Z^\prime}_r^{T})\widetilde{\mathcal{B}}_j
\end{split}
\end{equation}
where $\alpha_{i,j}$ represents the element in the $i$-th row and the $j$-th column.

After that, the bilinear matrix $\alpha$ is obtained by a softmax function over the $\rho\times\phi$ regions given by:
\begin{equation}
\setlength{\abovedisplayskip}{12pt}
\setlength{\belowdisplayskip}{12pt}
\alpha:= softmax(((\mathcal{I}\cdot {p}^T)\otimes\widetilde{\mathcal{D}}^{T}U)Z^{T}\widetilde{\mathcal{B}})   
\label{eq:11}              
\end{equation}
\begin{equation}
\alpha_{i,j} = p^{T}((U^{T}\widetilde{\mathcal{D}}_{i})\otimes(Z^{T}\widetilde{\mathcal{B}}_{j}))   
\label{eq:12}                
\end{equation}
where $p\in\mathbb{R}^{{\gamma}^\prime} $, and $\mathcal{I}\in\mathbb{R}^\rho$ is represented as an identity matrix.

Using low-rank bilinear pooling, the multi-head attention model $\alpha_g$ is shown in Eq.~(\ref{eq:13}). 
\begin{equation}
\setlength{\abovedisplayskip}{12pt}
\setlength{\belowdisplayskip}{12pt}
\alpha_g := softmax(((\mathcal{I}\cdot {p_{g}}^T)\otimes \widetilde{\mathcal{D}}^{T}U)Z^{T}\widetilde{\mathcal{B}})
\label{eq:13}      
\end{equation}
in which $g$ indicates the index of glimpses. It is worth noting that $p^T$, $U^T$, and $Z^T$ are denoted as trainable parameters. The difference between Eq.~(\ref{eq:13}) and Eq.~(\ref{eq:11}) is the initialization value of $p^T$, which affects the later iterative training results. Finally, different attention parameters can be trained to achieve a multi-head attention model as below:
\begin{equation}
\setlength{\abovedisplayskip}{12pt}
\setlength{\belowdisplayskip}{12pt}
\mathcal{O}_{intra} = Att_{intra}(\mathcal{D},\mathcal{B};\alpha,W_1,W_2,b_1,b_2)  
\label{eq:14}                 
\end{equation}
where $Att_{intra}(\cdot)$ simulates the whole intra-modal attention mechanism to obtain $\mathcal{O}_{intra}$.

\textit{2) Inter-modality relation Attention:}
Given an image $\mathcal{V}$, the pre-trained LOIS is employed to extract the features of detection regions as $v= \{v_1, v_2, \dots, v_l\}, v\in  \mathbb{R}^{m\times1}$. The word vectors are fed into a BERT model to extract the sequence as the question representation. We set $q= \{q_1, q_2,\dots, q_l\}$, $q_i\in \mathbb{R}^{n\times h}$ as question $\mathcal{Q}$, which indicates the important output of the $i$-th word in the question. Here, $m$ and $n$ denote the different dimensionalities of the feature vector, respectively. The length of $h$ is set to 14 after that. 

Similar to intra-modal attention, a bilinear model for the two groups of input channels is listed as follows:
\begin{equation}
\setlength{\abovedisplayskip}{12pt}
\setlength{\belowdisplayskip}{12pt}
\widetilde{v}=\tanh(p^{T}W^{\prime}_3+b_3), W^{\prime}_3\in\mathbb{R}^{min(m,n)\times 1}, b_3\in\mathbb{R}^1      
\label{eq:15} 
\vspace*{5pt}     
\end{equation}
\begin{equation}
\widetilde{q}=\tanh(q^{T}W^{\prime}_4+b_4), W^{\prime}_4\in\mathbb{R}^{min(m,n)\times h}, b_4\in\mathbb{R}^h      
\label{eq:16} 
\vspace*{5pt}     
\end{equation}
where $v\in\mathbb{R}^{m\times1}$, $q\in\mathbb{R}^{n\times h}$, $m=len(\mathcal{O}_{intra})$, and $h=\mid\{q_l\}\mid$. $min(m,n)$ is represented as image and question features compressed into shorter vector dimensions, respectively. $W^{\prime}_3$, $W^{\prime}_4$, $b_3$ and $b_4$ are both trainable parameters. Here, we mainly pay attention to two cases. One is that the vector lengths of the two modalities need to be aligned. The second is to reduce computational complexity and improve efficiency.

A bilinear attention map is calculated with the following formula:
\vspace*{5pt}
\begin{equation}
\setlength{\abovedisplayskip}{15pt}
\setlength{\belowdisplayskip}{15pt}
\mathcal{O}^{\prime}_{inter,r} = (\widetilde{v}^{T}U^{\prime})^{T}_{r} \beta(\widetilde{q}^{T}Z^{\prime})_r 
\label{eq:17}  
\vspace*{5pt} 
\end{equation}
where $\widetilde{v}\in\mathbb{R}^{{min(m,n)}\times{1}}$, $\widetilde{q}\in\mathbb{R}^{{min(m,n)}\times{\phi}}$, and $U^{\prime}\in\mathbb{R}^{{min(m,n)}\times{\gamma}}$, $Z^{\prime}\in\mathbb{R}^{{min(m,n)}\times{\gamma}}$. Then, the bilinear matrix $\beta$ can be expressed as:
\begin{equation}
\setlength{\abovedisplayskip}{12pt}
\setlength{\belowdisplayskip}{12pt}
\beta := softmax(((\mathcal{I}\cdot p^T )\otimes \widetilde{v}^{T}U)Z^{T} \widetilde{q}),  \beta\in\mathbb{R}^{1\times h}              
\label{eq:18} 
\end{equation}
\begin{equation}
\beta_{i,j} = p^T ((U^{T}\widetilde{v}_{i})\otimes(Z^{T}\widetilde{q}_{j}))                    
\label{eq:19} 
\vspace*{1pt} 
\end{equation}
where $\mathcal{I}\in\mathbb{R}^1$. Meanwhile, the multi-head attention model $\beta_g$ is defined as follows:
\begin{equation}
\setlength{\abovedisplayskip}{12pt}
\setlength{\belowdisplayskip}{12pt}
\beta_{g} := softmax(((\mathcal{I}\cdot p_{g}^T)\otimes\widetilde{v}^{T}U)Z^{T}\widetilde{q})                 
\label{eq:20} 
\end{equation}

The whole inter-modality operation from the image and question features ($v$, $q$) can be formalized as:
\begin{equation}
\setlength{\abovedisplayskip}{12pt}
\setlength{\belowdisplayskip}{12pt}
\mathcal{O}_{inter}= Att_{inter}(v,q;\beta,W_{3},W_{4},b_{3},b_{4})       
\label{eq:21}       
\end{equation}

As calculated above, we give a set $A= \{A_1, A_2, \dots, A_T\}$ as the answer vector of length $T$ and then feed it to the output classifier.
\begin{equation}
\setlength{\abovedisplayskip}{12pt}
\setlength{\belowdisplayskip}{12pt}
A_{T}=Att_{inter}(Att_{intra}(\mathcal{D},\mathcal{B};\alpha),q;\beta)\mid_{W,b}    
\label{eq:22}                 
\end{equation}
where $W=\{W_1,W_2,W_3,W_4\}$ and $b=\{b_1,b_2,b_3,b_4\}$ are the full connected weights and biases, respectively. Finally, to update each visual region and word feature, information is transferred through the learned weights and aggregate features between the two modalities.  

\section{Experiments}
In this section, some setups are introduced in our experiments. We train our model on four benchmark datasets to evaluate the performance of the developed LOIS. Moreover, ablation experiments are also implemented extensively on the dataset to explore the effect of different hyper-parameters. Finally, the qualitative results are provided for comparison with current state-of-the-art methods.

\subsection{Experimental Settings}
Our model initializes SOLO v2 with an underlying Resnet-101 backbone to fuse Feature Pyramid Network (FPN), which is used to pre-train annotations from ImageNet (classification). Then, we train the feature extractor on Microsoft COCO. The predefined grid numbers $S=[12, 16, 24, 36, 40]$ are used herein to improve model capability. A confidence $score\_threshold$ is set to 0.1 for filtering out predictions with low confidence. For each predicted mask, we calculate accurate instance masks by setting the $mask\_threshold$ to 0.5. The dimensionality of the semantic category is set to 80. Input visual features $\mathbb{R}^{(\rho+\phi)\times2048}$ are extracted from the proposed LOIS framework without bounding boxes, where $\rho+\phi\in[10, 100]$ is the number of detected objects with a confidence threshold. For the word features, we pad all questions with 0 to a maximum length of 14 and extract $\mathbb{R}^{14\times768}$ as question embedding $\mathcal{Q}$ by BERT. The BERT model is pre-trained on Wikipedia (2500M) and BooksCorpus (800M words). After fusing visual and word features, we transform them into 1024 dimensions by a fully-connected layer. The number of multi-head attentions is set to 8. For the optimization model, a dropout rate of 0.5 is used for all fully-connected layers. All gradients are clipped to 0.25, and the batch size is fixed at 256. All models are trained for 100 epochs with an initial learning rate of $lr=1\times10^{-2}$, and the warm-up ratio is set to 1/3. Stochastic gradient descent (SGD) is selected as an optimizer with a momentum of 0.9 and a weight decay of $10^{-4}$. Furthermore, a GAN-based network is employed to fill up pixels in the mask for image inpainting, which is pre-trained on multiple datasets, including faces (CelebA-HQ\footnote{https://github.com/tkarras/progressive\_growing\_of\_gans}) and natural images (Places2\footnote{http://places2.csail.mit.edu/}, ImageNet). Our operations are implemented by PyTorch 1.10.2 and Tensorflow 2.3.0. All the codes are run on four NVIDIA A100 GPUs. It costs 0.02 s for each image in training the grid cells. The mask features are computed with 0.11 s at each image. We use the LOIS to extract foreground instance objects at a speed of 0.19 s per image. Furthermore, we take 0.17 s for each step and a total of 10.1 h for 30 epochs during the whole training process.
\begin{table}[!t]
\centering
\renewcommand{\arraystretch}{1.3}
\setlength{\abovecaptionskip}{0.3cm}
\caption{Statistics of the different VQA datasets}
\label{tab_1}
\begin{IEEEeqnarraybox}[\IEEEeqnarraystrutmode\IEEEeqnarraystrutsizeadd{2pt}{2pt}]{l/c/c/c/c}
\IEEEeqnarrayrulerow[1pt]\\
\mbox{Dataset} ~& \mbox{VQA v1} ~~& \mbox{VQA v2} ~~&\mbox{COCO-QA} ~&\mbox{VQA-CP v2}\\
\IEEEeqnarrayrulerow[0.5pt]\\
\raisebox{-7pt}[0pt][0pt]{Split}  ~~& \mbox{train/val} ~& \mbox{train/val} ~~& \mbox{train}   ~& \mbox{train}\\
~& \mbox{test-dev/test-std} ~~&  \mbox{test-dev/test-std} ~~&\mbox{test-std} ~&\mbox{test-std}\\
\IEEEeqnarrayrulerow[1pt]\\
\end{IEEEeqnarraybox}
\vspace*{-17pt}
\end{table}   
\subsection{Datasets and Evaluation Metric}
\emph{VQA v1}\footnote{https://visualqa.org/vqa\_v1\_download.html} is the first widely employed benchmark for the VQA dataset. It contains 248,349 questions for training, 244,302 questions for testing, and 121,512 questions for validation. Here, 204,721 natural images are obtained from the Microsoft COCO dataset. Additionally, the testing set is divided into test-std and test-dev. The question-answer pairs in VQA v1 are annotated based on human annotators and contain three types of corresponding answers, which are “Yes/No”,  “Number”, and  “Other”. The VQA v1 dataset consists of open-ended and multiple-choice two subtasks.

\emph{VQA v2}\footnote{https://visualqa.org/download.html} is an extended version of VQA v1, which focuses on reducing dataset biases through balanced pairs. The whole dataset consists of 443,757 training questions, 214,354 validation questions, and 447,793 test-std questions. Among the test-std tests, 25\% subset exists as test-dev. Each question generates 10 free-response answers from human annotators. An accuracy-based evaluation metric is provided to predict answer $A$ by \cite{b1} as follows:

\begin{equation}
Accuracy(A)=min\left(\frac{\#humans\ that\ said\ (A)}{3}, 1\right)
\label{eq:23}   
\end{equation}\\
where $A$ is the number of provided answers by different annotators. If the predicted answer is given by more than three annotators, the corresponding score is “1”.

\emph{COCO-QA}\footnote{https://cocodataset.org/} is created from the Microsoft COCO dataset, which is smaller than VQA v1 and VQA v2. This dataset contains 66.9\% training and 33.1\% test sets. The whole dataset consists of 69,172 images, 92,396 questions, and 435 answers. The questions are generated in four types: “Object (69.84\%)”, “Number (16.59\%)”,  “Color (7.47\%)” and  “Location (6.10\%)”. In addition, Wu-Palmer similarity (WUPS) is adopted as an additional metric to measure the semantic distance between the ground-truth answers and the predicted ones. The WUPS scores are set to 0.0 and 0.9.

\emph{VQA-CP v2}\footnote{https://computing.ece.vt.edu/$\sim$aish/vqacp/} is reorganized from VQA v2, which has changed the prior distributions of answers to reduce the question-oriented bias in train and test splits. Specifically, it contains 121$k$ images, 438$k$ questions, and 4.4$M$ answers for the training set. The test set of VQA-CP v2 has 98$k$ images, 220$k$ questions, and 2.2$M$ answers.
\begin{table*}[!t]
\centering
\renewcommand{\arraystretch}{1.3}
\setlength{\abovecaptionskip}{0.3cm}
\caption{Effectiveness of the image features on the VQA v1, VQA v2, and VQA-CP v2 datasets. The best results are in bold. \\Note that “LOIS” is the method without GAN. And “LOIS+GAN” is the method of using GAN in this paper.}
\label{tab_2}
\begin{IEEEeqnarraybox}[\IEEEeqnarraystrutmode\IEEEeqnarraystrutsizeadd{1pt}{1pt}]{l/c/c/c/c/c/c/c}
\IEEEeqnarrayrulerow[1pt]\\
\raisebox{-7pt}[0pt][0pt]{Dataset}~~~~~~~&\raisebox{-8pt}[0pt][0pt]{Method}~~~~~~~&\IEEEeqnarraymulticol{2}{t}{VQA v1} &\IEEEeqnarraymulticol{2}{t}{VQA v2} &\IEEEeqnarraymulticol{2}{t}{VQA-CP v2} \\
\cmidrule[0.5pt]{3-8}
&  &\mbox{LOIS}~~~~~~~&\mbox{LOIS+GAN}~~~~~~~&\mbox{LOIS}~~~~~~~&\mbox{LOIS+GAN}~~~~~~~&\mbox{LOIS}~~~~~~~&\mbox{LOIS+GAN}\\         
\IEEEeqnarrayrulerow[0.5pt]\\
 \raisebox{-15pt}[0pt][0pt]{Test-dev (\%)}~~~~~~~&\mbox{Y/N} ~~~~~~~&75.01~~~~~~~&87.72~~~~~~~&75.57~~~~~~~&87.91~~~~~~~&-~~~~~~~&-\\
&\mbox{Num} ~~~~~~~&29.69~~~~~~~&46.24~~~~~~~&34.68~~~~~~~&54.34~~~~~~~&-~~~~~~~&-\\
&\mbox{Other} ~~~~~~~&39.87~~~~~~~&61.45~~~~~~~&40.99~~~~~~~&61.45~~~~~~~&-~~~~~~~&-\\
 &\mbox{Overall} ~~~~~~~&52.02~~~~~~~&\textbf{71.11}~~~~~~~&54.13~~~~~~~&\textbf{72.78}~~~~~~~&-~~~~~~~&-\\
\IEEEeqnarrayrulerow[0.5pt]\\
 \raisebox{-15pt}[0pt][0pt]{Test-std (\%)}~~~~~~~&\mbox{Y/N} ~~~~~~~&75.22~~~~~~~&87.99~~~~~~~&72.96~~~~~~~&88.99~~~~~~~&40.55~~~~~~~&51.82 \\
&\mbox{Num} ~~~~~~~&29.91~~~~~~~&46.24~~~~~~~&36.05~~~~~~~&55.08~~~~~~~&11.05~~~~~~~&14.24 \\
&\mbox{Other} ~~~~~~~&40.26~~~~~~~&61.37~~~~~~~&40.95~~~~~~~&66.37~~~~~~~&30.79~~~~~~~&41.39 \\
&\mbox{Overall} ~~~~~~~&53.95~~~~~~~&\textbf{70.68}~~~~~~~&54.33~~~~~~~&\textbf{73.02}~~~~~~~&29.32~~~~~~~&\textbf{43.09} \\
\IEEEeqnarrayrulerow[1pt]\\
\end{IEEEeqnarraybox}
\vspace*{-12pt}
\end{table*}    
\begin{table*}[!t]
\centering
\renewcommand{\arraystretch}{1.3}
\setlength{\abovecaptionskip}{0.3cm}
\caption{Ablation study for different sequences in attention embedding on the VQA v2 datasets.
\\Note that “I-B-Q” is our selected model in this work. The best results are in bold.}
\label{tab_3}
\begin{IEEEeqnarraybox}[\IEEEeqnarraystrutmode\IEEEeqnarraystrutsizeadd{1pt}{1pt}]{l/c/c/c/c/c/c/c/c/c/c/c}
\IEEEeqnarrayrulerow[1pt]\\
\raisebox{-8pt}[0pt][0pt]{Method}~~~~~~&&\IEEEeqnarraymulticol{4}{t}{Test-dev (\%)}&&\IEEEeqnarraymulticol{5}{t}{Test-std (\%)}\\
\cmidrule[0.5pt]{3-6}\cmidrule[0.5pt]{8-12}
&&\mbox{Y/N}&~~~~~~~~\mbox{Num}&~~~~~~~~\mbox{Other}&~~~~~~~\mbox{Overall}&~~~~~&\mbox{Y/N}&~~~~~~~~\mbox{Num}&~~~~~~~~\mbox{Other}&~~~~~~~\mbox{Overall} \\         
\IEEEeqnarrayrulerow[0.5pt]\\
\mbox{No Attention}~~~~~~&&81.78&~~~~~~~~47.14&~~~~~~~~50.24&~~~~~~~~64.52&&81.85&~~~~~~~~45.73&~~~~~~~~53.01&~~~~~~~63.01\\
\mbox{B-Q} ~~~~~~&&82.43&~~~~~~~~47.69&~~~~~~~~50.89&~~~~~~~~65.35&&82.96 &~~~~~~~~46.28&~~~~~~~~53.95&~~~~~~~64.23\\
\mbox{I-B} ~~~~~~&&82.99&~~~~~~~~49.41&~~~~~~~~57.34&~~~~~~~~67.83&&81.82 &~~~~~~~~49.29&~~~~~~~~56.99&~~~~~~~66.38\\
\mbox{I-Q}~~~~~~&&84.68&~~~~~~~~49.74&~~~~~~~~59.31&~~~~~~~~69.03&&82.33 &~~~~~~~~49.90&~~~~~~~~58.96&~~~~~~~69.67\\
\mbox{B-Q-I}~~~~~~&&86.66&~~~~~~~~52.10&~~~~~~~~58.97&~~~~~~~~69.92&&86.85 &~~~~~~~~53.42&~~~~~~~~63.85&~~~~~~~70.98\\
\mbox{I-Q-B}~~~~~~&&86.90&~~~~~~~~53.21&~~~~~~~~59.91&~~~~~~~~70.33&&87.13 &~~~~~~~~54.00&~~~~~~~~64.94&~~~~~~~71.19\\
\mbox{\textbf{I-B-Q (ours)}}~~~~~~&&\textbf{87.91}&~~~~~~~~\textbf{54.34}&~~~~~~~~\textbf{61.45}&~~~~~~~~\textbf{72.78}&&\textbf{88.99}&~~~~~~~~\textbf{55.08}&~~~~~~~~\textbf{66.37}&~~~~~~~\textbf{73.02}\\
\IEEEeqnarrayrulerow[1pt]\\
\end{IEEEeqnarraybox}
\vspace*{-16pt}
\end{table*}  
\begin{figure}[!t]
\centering
\subfloat{\includegraphics[width=1.73in]{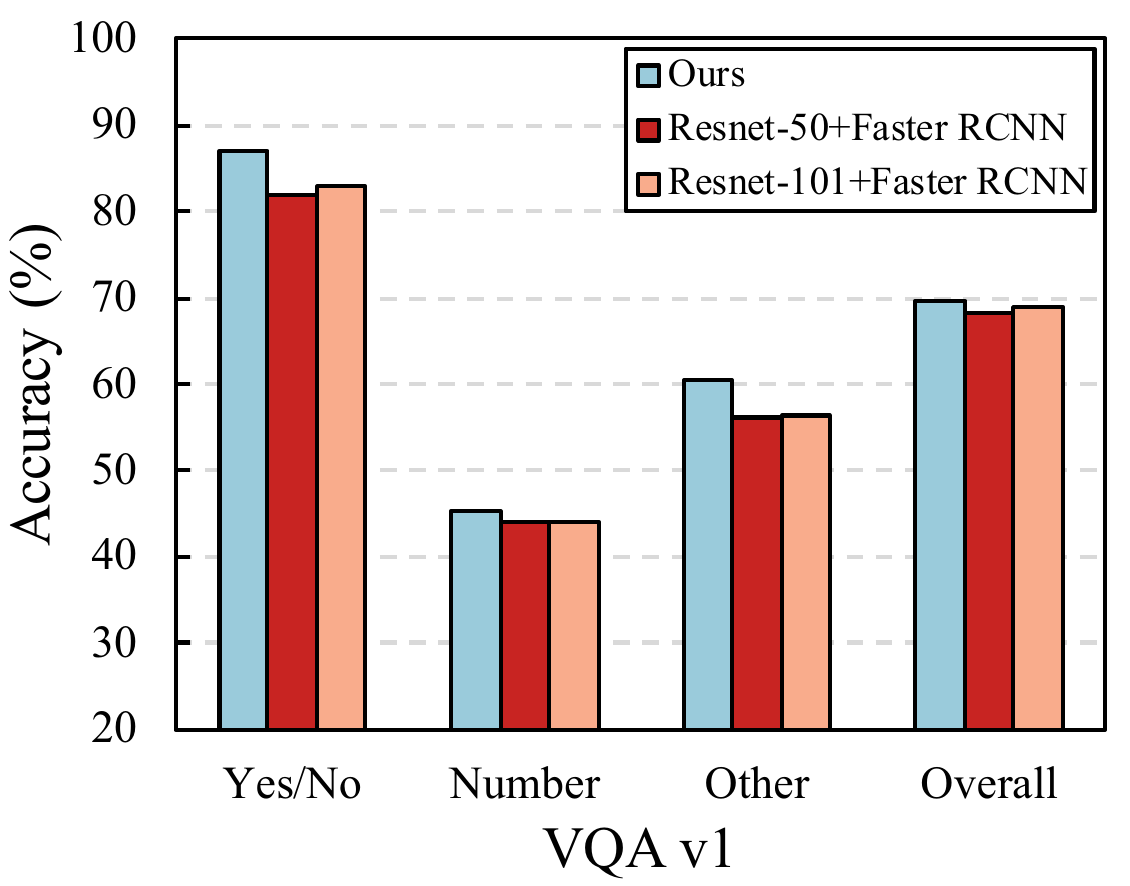}}
\hfil
\subfloat{\includegraphics[width=1.73in]{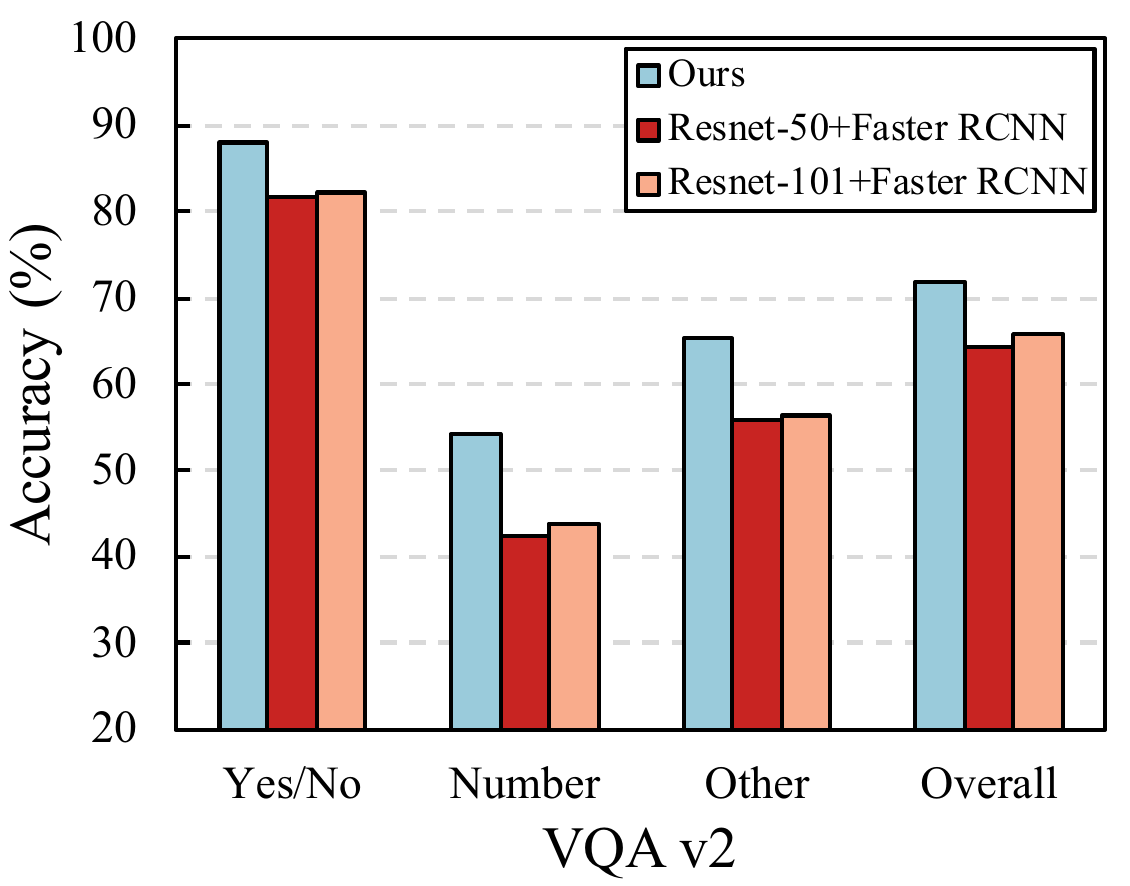}}
\hfil
\subfloat{\includegraphics[width=1.73in]{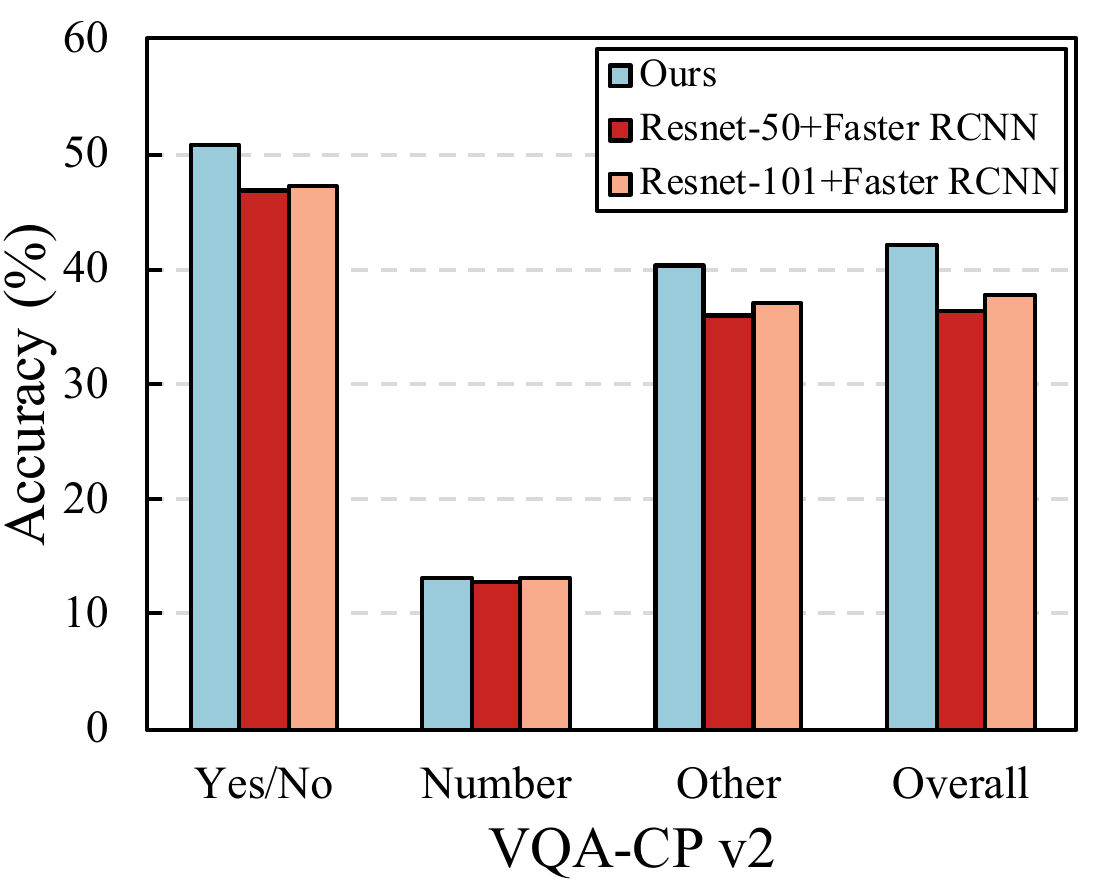}}
\hfil
\subfloat{\includegraphics[width=1.73in]{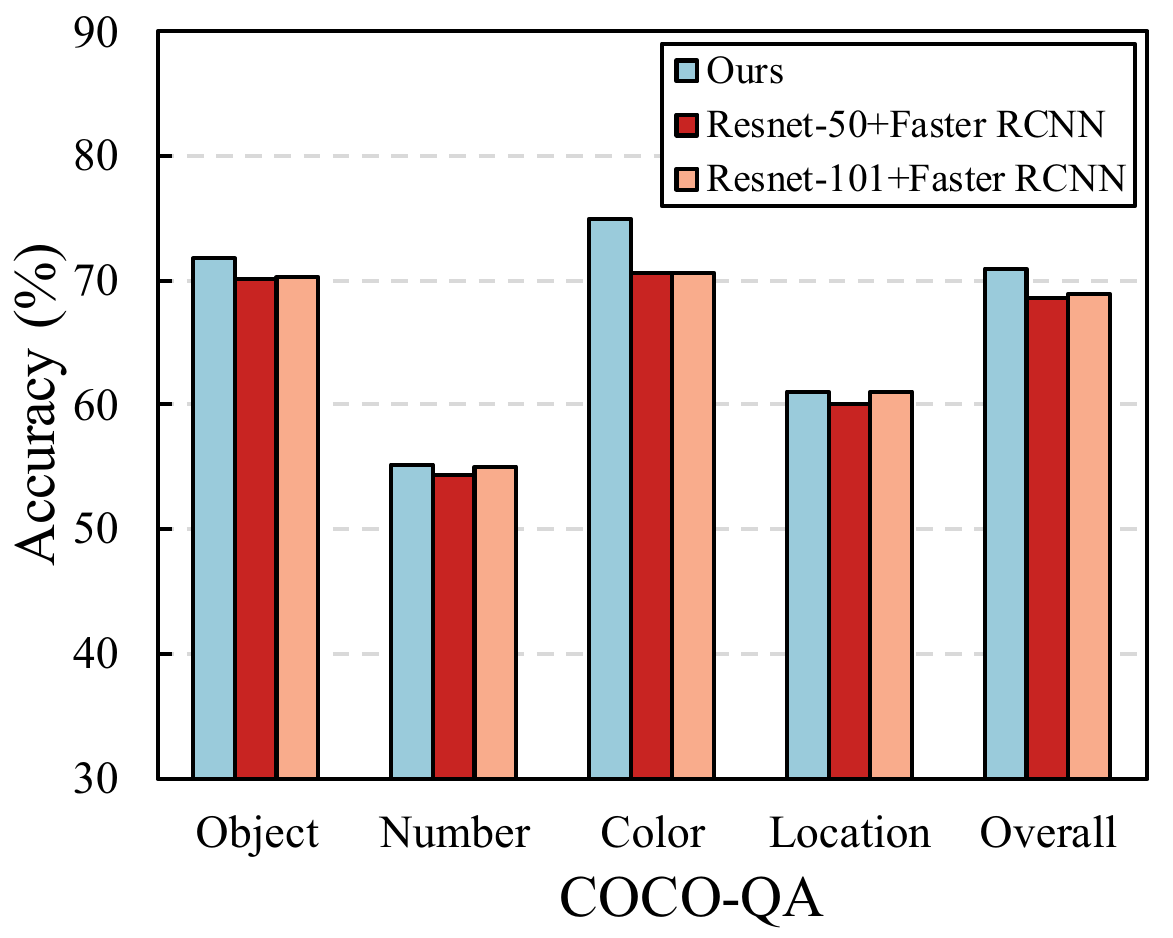}}
\caption{The accuracy of the mainstream Faster RCNN features and our method under different categories.} \label{fig_7}
\vspace*{-6pt}
\end{figure}
\subsection{Ablation Study of LOIS}
We further perform ablation studies to validate the effectiveness of each part in our designed LOIS. Be aware that all the models have the same input for a fair comparison. In Table~\ref{tab_1}, all the evaluated models are trained on the VQA v1, VQA v2, COCO-QA, and VQA-CP v2 datasets.

First, the superiority with and without the GAN is compared in our model. “LOIS (No GAN)” indicates the model without GAN, which directly carries out a uniform grid method to model the remaining background features after extracting the instance masks. “LOIS” indicates the proposed model with GAN. Here, a GAN network is used to fill up the pixels in the background mask to prevent clues about the shape or outline of the missing object. As shown in Table~\ref{tab_2}, the model without GAN has a decrease in accuracy due to the loss of background features. Thus, “LOIS+GAN” is chosen as our model in this work. In addition, we present experimental results compared with the current mainstream Faster RCNN features. “Resnet-50” and “Resnet-101” provide a comparison of the performance at different depths. To exhibit that our method can improve the accuracy of each component, the performance on the different dataset splits is shown in Fig.~\ref{fig_7}. A slight improvement is observed in the accuracy of the Resnet-101 backbone. In particular, we find significant improvements in our model on the VQA v2 and VQA-CP v2 datasets. The main reason is that image features can be better fused on these datasets, both of which are tuned for the dataset distribution. It further helps us to enhance the detection of image semantics. 
\begin{figure}[!t]
\centering
\includegraphics[width=3in]{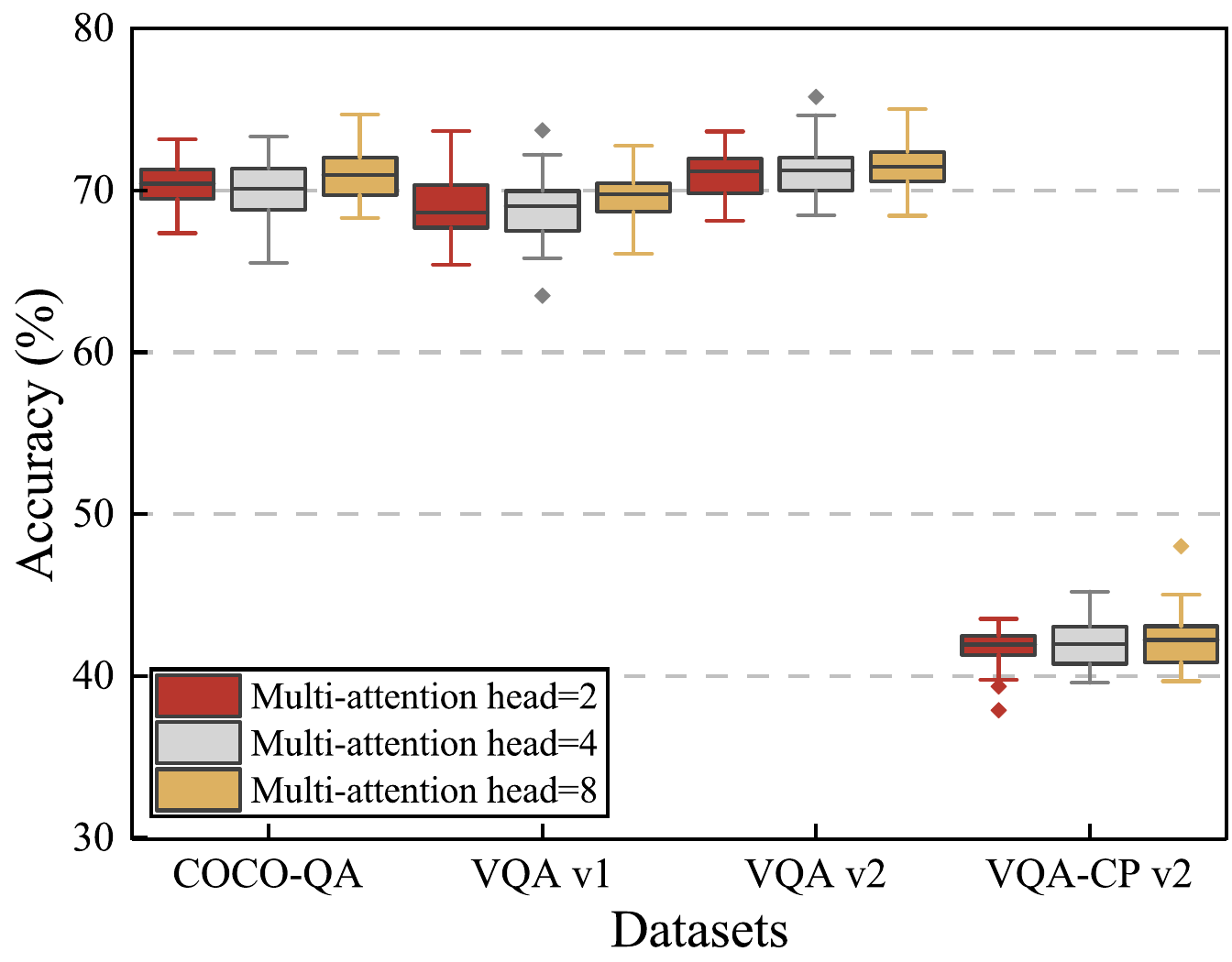}
\caption{Experimental results of the multi-head attention mechanism at head=2, 4, and 8.}
\label{fig_8}
\vspace*{-20pt}
\end{figure}
\begin{table*}[!t]
\centering
\renewcommand{\arraystretch}{1.3}
\setlength{\abovecaptionskip}{0.3cm}
\caption{Performance evaluation of feature extraction module on the VQA v2 dataset. \\The best results are in bold.}
\label{tab_4}
\begin{IEEEeqnarraybox}[\IEEEeqnarraystrutmode\IEEEeqnarraystrutsizeadd{1pt}{1pt}]{l/c/c/c/c/c/c/c/c/c/c/c/c}
\IEEEeqnarrayrulerow[1pt]\\
\raisebox{-9pt}[0pt][0pt]{Method}~~&\raisebox{-9pt}[0pt][0pt]{Faster RCNN}~~~&\IEEEeqnarraymulticol{4}{t}{\mbox{Test-dev (\%)}}&&\IEEEeqnarraymulticol{5}{t}{\mbox{Test-std (\%)}}\\
\cmidrule[0.5pt]{3-6}\cmidrule[0.5pt]{8-13}
&&\mbox{Y/N}~~~~~~&\mbox{Num}~~~~~~&\mbox{Other}~~~~~~&\mbox{Overall} &~~~&\mbox{Y/N}~~~~~~&\mbox{Num}~~~~~~&\mbox{Other}~~~~~~&\mbox{Overall} \\  
\IEEEeqnarrayrulerow[0.5pt]\\
\mbox{Background}~~&\usym{1F5F4}~~~~&45.34~~~~~~&21.23~~~~~~&26.54~~~~~~&36.19&~~~&46.01~~~~~~&20.55~~~~~~&26.95~~~~~~&36.33\\                                                                            
\mbox{Foreground}~~&\usym{1F5F4}~~~~&75.78~~~~~~&44.01~~~~~~&51.27~~~~~~&63.51&~~~&75.82~~~~~~&40.29~~~~~~&50.98~~~~~~&62.98\\    
\mbox{Background+Grid}~~&\usym{1F5F4}~~~~&80.85~~~~~~&45.94~~~~~~&53.35~~~~~~&66.97&~~~&79.25~~~~~~&43.90~~~~~~&52.28~~~~~~&65.02\\
\mbox{Background+Faster RCNN}~~&\usym{2713}~~~~&84.90~~~~~~&52.97~~~~~~&58.88~~~~~~&69.93&~~~&85.11~~~~~~&53.08~~~~~~&63.29~~~~~~&69.22\\ 
\IEEEeqnarrayrulerow[0.5pt]\\
\IEEEeqnarrayseprow[2pt]\\
\mbox{\textbf{LOIS (ours)}}~~&\usym{1F5F4}~~~~&\textbf{87.91}~~~~~~&\textbf{54.34}~~~~~~&\textbf{61.45}~~~~~~&\textbf{72.78}&~~~&\textbf{88.99}~~~~~~&\textbf{55.08}~~~~~~&\textbf{66.37}~~~~~~&\textbf{73.02}\\
\IEEEeqnarrayrulerow[1pt]\\
\end{IEEEeqnarraybox}
\vspace*{-7pt}
\end{table*} 

\begin{table*}[!t]
\centering
\renewcommand{\arraystretch}{1.3}
\setlength{\abovecaptionskip}{0.3cm}
\caption{Performance evaluation of LOIS and comparative methods on the VQA v1 dataset. \\Bold values highlight the best experimental results for the comparison of the methods.}
\label{tab_5}
\begin{IEEEeqnarraybox}[\IEEEeqnarraystrutmode\IEEEeqnarraystrutsizeadd{1pt}{1pt}]{l/c/c/c/c/c/c/c/c/c/c/c/c}
\IEEEeqnarrayrulerow[1pt]\\
\raisebox{-9pt}[0pt][0pt]{Method}~~~~~&\raisebox{-9pt}[0pt][0pt]{Faster RCNN}~~~~~&\IEEEeqnarraymulticol{4}{t}{\mbox{Test-dev (\%)}}&&\IEEEeqnarraymulticol{5}{t}{\mbox{Test-std (\%)}}\\
\cmidrule[0.5pt]{3-6}\cmidrule[0.5pt]{8-13}
&&\mbox{Y/N}~~~~~~~&\mbox{Num}~~~~~~~&\mbox{Other}~~~~~~~&\mbox{Overall} &~~~~~&\mbox{Y/N}~~~~~~~&\mbox{Num}~~~~~~~&\mbox{Other}~~~~~~~&\mbox{Overall} \\  
\IEEEeqnarrayrulerow[0.5pt]\\
\mbox{SAN}\mbox{\cite{b35}}~~~~~&\usym{1F5F4}~~~~~~&79.30~~~~~~~&36.60~~~~~~~&46.10~~~~~~~&58.70&~~~~~&79.11~~~~~~~&36.41~~~~~~~&46.42~~~~~~~&58.85\\                                                               
\mbox{Dual-MFA}\mbox{\cite{b25}}~~~~~&\usym{2713}~~~~~~&83.59~~~~~~~&40.18~~~~~~~&56.84~~~~~~~&66.01&~~~~~&83.37~~~~~~~&40.39~~~~~~~&56.89~~~~~~~&66.09\\             
\mbox{MLB}\mbox{\cite{b45}}~~~~~&\usym{1F5F4}~~~~~~&85.57~~~~~~~&39.32~~~~~~~&57.36~~~~~~~&67.03&~~~~~&84.39~~~~~~~&38.70~~~~~~~&58.20~~~~~~~&66.96\\    
\mbox{ODA}\mbox{\cite{b33}}~~~~~&\usym{2713}~~~~~~&85.82~~~~~~~&43.03~~~~~~~&58.07~~~~~~~&67.83&~~~~~&85.81~~~~~~~&42.51~~~~~~~&58.24~~~~~~~&67.97\\
\mbox{DA-NTN}\mbox{\cite{b46}}~~~~~&\usym{2713}~~~~~~&85.80~~~~~~~&41.90~~~~~~~&58.60~~~~~~~&67.90&~~~~~&85.80~~~~~~~&42.50~~~~~~~&58.50~~~~~~~&68.10\\ 
\mbox{CoR-3}\mbox{\cite{b47}}~~~~~&\usym{2713}~~~~~~&85.69~~~~~~~&44.06~~~~~~~&59.08~~~~~~~&68.37&~~~~~&85.83~~~~~~~&43.93~~~~~~~&59.11~~~~~~~&68.54\\ 
\mbox{ALMA}\mbox{\cite{b48}}~~~~~&\usym{1F5F4}~~~~~~&85.49~~~~~~~&42.09~~~~~~~&59.97~~~~~~~&68.94&~~~~~&84.11~~~~~~~&42.59~~~~~~~&58.06~~~~~~~&68.76\\ 
\mbox{DenIII}\mbox{\cite{b49}}~~~~~&\usym{2713}~~~~~~&86.70~~~~~~~&44.10~~~~~~~&59.70~~~~~~~&69.10&~~~~~&86.80~~~~~~~&43.30~~~~~~~&59.40~~~~~~~&69.00\\ 
\mbox{CAM}\mbox{\cite{b50}}~~~~~&\usym{2713}~~~~~~&86.77~~~~~~~&42.78~~~~~~~&60.27~~~~~~~&69.25&~~~~~&86.55~~~~~~~&42.36~~~~~~~&60.38~~~~~~~&69.29\\
\mbox{MRA-Net}\mbox{\cite{b41}}~~~~~&\usym{2713}~~~~~~&86.79~~~~~~~&43.89~~~~~~~&59.62~~~~~~~&69.06&~~~~~&86.37~~~~~~~&44.16~~~~~~~&60.00~~~~~~~&69.22\\
\mbox{ALSA}\mbox{\cite{b39}}~~~~~&\usym{2713}~~~~~~&87.12~~~~~~~&42.94~~~~~~~&59.06~~~~~~~&69.52&~~~~~&86.94~~~~~~~&43.84~~~~~~~&58.21~~~~~~~&69.32\\
\IEEEeqnarrayrulerow[0.5pt]\\
\IEEEeqnarrayseprow[2pt]\\
\mbox{\textbf{LOIS (ours)}}~~~~~&\usym{1F5F4}~~~~~~&\textbf{87.72}~~~~~~~&\textbf{46.24}~~~~~~~&\textbf{61.45}~~~~~~~&\textbf{71.11}&~~~~~&\textbf{87.99}~~~~~~~&\textbf{46.24}~~~~~~~&\textbf{61.37}~~~~~~~&\textbf{70.68}\\
\IEEEeqnarrayrulerow[1pt]\\
\end{IEEEeqnarraybox}
\vspace*{-11pt}
\end{table*} 

Next, we report an experiment to verify the effect of attention embedding by different operation sequences. We mainly carry out analysis from the following two aspects: 1) Suppose that attention is used only once, while the other attention is replaced by matrix multiplication. We provide three cases, “I-B”, “I-Q” and “B-Q”, respectively. Specifically, “I-B” means an attention model applied to instances and background features. “I-Q” refers to the attention model applied to instances and questions. The “B-Q” indicates that the same method is performed for background features and questions. 2) Different operation sequences for two types of attention. There are three cases, namely “B-Q-I”, “I-Q-B” and “I-B-Q”. “B-Q-I” means that the attention mechanism is first applied to the background and questions, and the second attention mechanism is used by combining the obtained vectors and instance objects. The “I-Q-B” and “I-B-Q” are the same as the method “B-Q-I” in Table~\ref{tab_3}. As can be seen from Table~\ref{tab_3}, “I-Q” achieves higher accuracy than “I-B” and “B-Q”, which means that the use of one attention between the images and questions is still not negligible. For “I-Q-B” and “B-Q-I”, it is more difficult to train two cross-model tasks. Conversely, “I-B-Q” is easier to learn due to the similar distribution within the modalities. Apparently, “I-B-Q” shows better performance. This fully demonstrates that, more than “I-Q”, the background  features “B” do matter in our model. It is more reasonable to use “I-B-Q” than others as our original model. 

We also evaluate the performance of the feature extraction module in Table~\ref{tab_4}. The first method “Background” requires only retaining background information, while removing the instance features extraction process. On the contrary, the method “Foreground” only employs the instance feature module, removing background information. Clearly, we can be seen that the results perform poorly, which indicates the necessity of both instance and background feature extraction. In addition, the “Background+Grid” adopts a uniform grid-based method to extract features, while preserving background information. Prediction results are lower than our LOIS, which reflects the importance of the one-to-one mapping between instance masks and semantic features. The method “Background+Faster RCNN” uses a bounding box-based detection method, which results are still lower than ours. This also shows that LOIS can effectively avoid the impact of rough edge features and noise extracted based on the bounding box.

\begin{figure}[!t]
\centering
\includegraphics[width=3in]{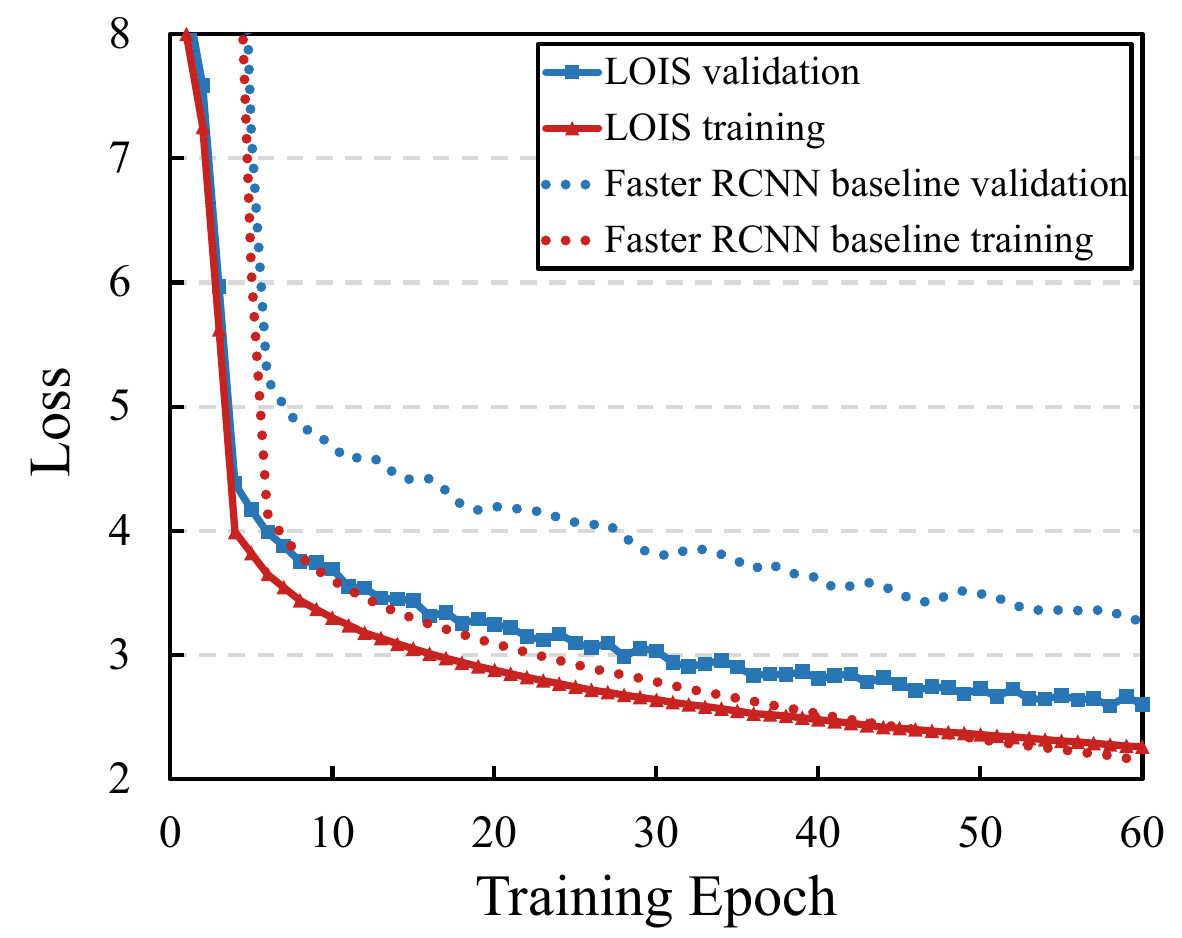}
\caption{Convergence of the proposed LOIS and Faster RCNN baselines from 1 to 60 epochs.}
\label{fig_9}
\vspace*{-8pt}
\end{figure}

\begin{figure}[!t]
\centering
\includegraphics[width=3in]{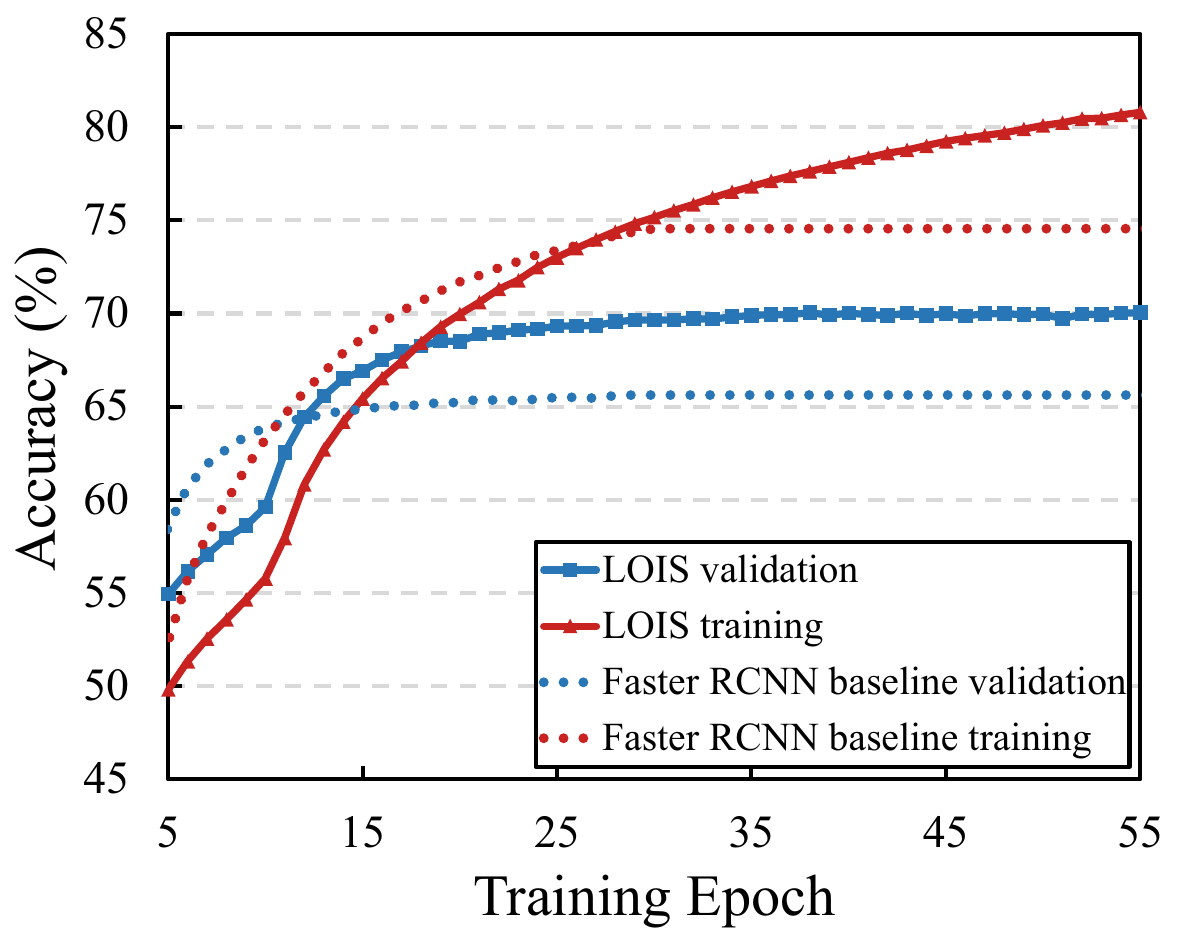}
\caption{The accuracy of the LOIS and Faster RCNN baselines on the VQA v2 dataset.}
\label{fig_10}
\vspace*{-3.5pt}
\end{figure}

In Fig.~\ref{fig_8}, the effect of multi-head attention on different datasets is analyzed. We set the hyper-parameters head=2, 4, and 8, respectively. As can be expected that the accuracy of the model increases with the number of heads. Therefore, we choose head=8 by experiments. Fig.~\ref{fig_9} records the loss values, which decrease rapidly with a certain vibration in the first 20 epochs. Compared with the Faster RCNN baseline, the LOIS model converges after about 60 epochs to fuse effective multimodal semantic features. Fig.~\ref{fig_10} displays the accuracy of the LOIS and Faster RCNN baseline. As shown on the validation set, LOIS is better than the Faster RCNN baseline after about 12 epochs. Meanwhile, after about 28 epochs, LOIS achieves higher accuracy on the training set, and experimental results verify the effectiveness of the LOIS, which provides a meaningful multi-view representation.
\begin{table*}[!t]
\centering
\renewcommand{\arraystretch}{1.3}
\setlength{\abovecaptionskip}{0.3cm}
\caption{Performance evaluation of LOIS and previous state-of-the-art methods on the VQA v2 dataset.}
\label{tab_6}
\begin{IEEEeqnarraybox}[\IEEEeqnarraystrutmode\IEEEeqnarraystrutsizeadd{1pt}{1pt}]{l/c/c/c/c/c/c/c/c/c/c/c/c}
\IEEEeqnarrayrulerow[1pt]\\
\raisebox{-10pt}[0pt][0pt]{Method}~~~&\raisebox{-10pt}[0pt][0pt]{Faster RCNN}~~~~~&\IEEEeqnarraymulticol{4}{t}{Test-dev (\%)}&&\IEEEeqnarraymulticol{6}{t}{Test-std (\%)}\\
\cmidrule[0.5pt]{3-6}\cmidrule[0.5pt]{8-13}
&&\mbox{Y/N}~~~~~~~&\mbox{Num}~~~~~~~&\mbox{Other}~~~~~~~&\mbox{Overall}&~~~~~&\mbox{Y/N}~~~~~~~&\mbox{Num}~~~~~~~&\mbox{Other}~~~~~~~&\mbox{Overall} \\  
\IEEEeqnarrayrulerow[0.5pt]\\
\mbox{MLB}\mbox{\cite{b45}}~~~&\usym{1F5F4}~~~~~&-~~~~~~~&-~~~~~~~&-~~~~~~~&-&~~~~~&84.02 ~~~~~~~&37.90~~~~~~~&54.77~~~~~~~&65.07\\
\mbox{Bottom-up}\mbox{\cite{b26}}~~~&\usym{2713}~~~~~&81.82~~~~~~~&44.21~~~~~~~&56.05~~~~~~~&65.32&~~~~~&82.20~~~~~~~&43.90~~~~~~~&56.26~~~~~~~&65.67\\
\mbox{CRA-Net}\mbox{\cite{b34}}~~~&\usym{2713}~~~~~&84.87~~~~~~~&49.46~~~~~~~&59.08~~~~~~~&68.61&~~~~~&85.21~~~~~~~&48.43~~~~~~~&59.42~~~~~~~&68.92\\
\mbox{DFAF}\mbox{\cite{b31}}~~~&\usym{2713}~~~~~&86.09~~~~~~~&53.32~~~~~~~&60.49~~~~~~~&70.22&~&- ~~~~~~~&- ~~~~~~~&- ~~~~~~~&70.34\\
\mbox{ViLBERT}\mbox{\cite{b51}}~~~&\usym{2713}~~~~~&-~~~~~~~&-~~~~~~~&-~~~~~~~&70.55&~~~~~&-~~~~~~~&-~~~~~~~&-~~~~~~~&70.92 \\
\mbox{VisualBERT}\mbox{\cite{b52}}~~~&\usym{2713}~~~~~&-~~~~~~~&-~~~~~~~&-~~~~~~~&70.80&~~~~~&-~~~~~~~&-~~~~~~~&-~~~~~~~&71.00\\
\mbox{VL-BERT}\mbox{\cite{b53}}~~~&\usym{2713}~~~~~&-~~~~~~~&-~~~~~~~&-~~~~~~~&71.79&~~~~~&-~~~~~~~&-~~~~~~~&-~~~~~~~&72.22\\
\mbox{LXMERT}\mbox{\cite{b54}}~~~&\usym{2713}~~~~~&-~~~~~~~&-~~~~~~~&-~~~~~~~&72.42&~~~~~&88.20~~~~~~~&54.20~~~~~~~&63.10~~~~~~~&72.50\\
\mbox{MCAN}\mbox{\cite{b55}}~~~&\usym{2713}~~~~~&86.82~~~~~~~&53.26~~~~~~~&60.72~~~~~~~&70.63&~&-~~~~~~~&-~~~~~~~&-~~~~~~~&70.90\\
\mbox{Grid}\mbox{\cite{b12}}~~~&\usym{1F5F4}~~~~~&-~~~~~~~&-~~~~~~~&-~~~~~~~&66.27&~&-~~~~~~~&-~~~~~~~&-~~~~~~~&-\\
\mbox{Pixel-BERT}\mbox{\cite{b13}}~~~&\usym{1F5F4}~~~~~&-~~~~~~~&-~~~~~~~&-~~~~~~~&71.35&~&-~~~~~~~&-~~~~~~~&-~~~~~~~&71.42\\
\mbox{UNITER}\mbox{\cite{b56}}~~~&\usym{2713}~~~~~&-~~~~~~~&-~~~~~~~&-~~~~~~~&72.70&~&-~~~~~~~&-~~~~~~~&-~~~~~~~&72.91\\
\mbox{ALMA}\mbox{\cite{b48}}~~~&\usym{1F5F4}~~~~~&-~~~~~~~&-~~~~~~~&-~~~~~~~&-&~~~~~&84.62~~~~~~~&47.08~~~~~~~&58.14~~~~~~~&68.12\\
\mbox{DenIII}\mbox{\cite{b49}}~~~&\usym{2713}~~~~~&85.60~~~~~~~&49.10~~~~~~~&60.60 ~~~~~~~&69.60&~~~~~&85.60~~~~~~~&49.00~~~~~~~&60.50~~~~~~~&69.70\\
\mbox{VL-T5}\mbox{\cite{b57}}~~~&\usym{2713}~~~~~&-~~~~~~~&-~~~~~~~&-~~~~~~~&-&~&-~~~~~~~&-~~~~~~~&-~~~~~~~&70.30\\
\mbox{CAM}\mbox{\cite{b50}}~~~&\usym{2713}~~~~~&85.18~~~~~~~&47.35~~~~~~~&59.76~~~~~~~&68.82 &~~~~~&85.22~~~~~~~&46.98~~~~~~~&59.91~~~~~~~&68.99\\
\mbox{MRA-Net}\mbox{\cite{b41}}~~~&\usym{2713}~~~~~&85.58~~~~~~~&48.92~~~~~~~&59.46~~~~~~~&69.02&~~~~~  &85.83~~~~~~~&49.22~~~~~~~&59.86~~~~~~~&69.46\\
\mbox{ALSA}\mbox{\cite{b39}}~~~&\usym{2713}~~~~~&85.73~~~~~~~&48.98~~~~~~~&59.17~~~~~~~&69.21&~~~~~
&- ~~~~~~~&- ~~~~~~~&- ~~~~~~~&-\\
\mbox{DAQC}\mbox{\cite{b58}}~~~&\usym{2713}~~~~~&82.15~~~~~~~&43.57~~~~~~~&56.39~~~~~~~&64.51&~~~~~
&- ~~~~~~~&- ~~~~~~~&- ~~~~~~~&-\\
\mbox{MLVQA}\mbox{\cite{b59}}~~~&\usym{2713}~~~~~&86.64~~~~~~~&51.90 ~~~~~~~&60.53~~~~~~~&70.30&~~~~~&-~~~~~~~&-~~~~~~~&-~~~~~~~&70.57\\
\IEEEeqnarrayrulerow[0.5pt]\\
\IEEEeqnarrayseprow[2pt]\\
\mbox{\textbf{LOIS (ours)}}~~~&\usym{1F5F4}~~~~~&\textbf{87.91}~~~~~~~&\textbf{54.34} ~~~~~~~&\textbf{61.45}~~~~~~~&\textbf{72.78}&~~~~~& \textbf{88.99}~~~~~~~&\textbf{55.08}~~~~~~~&\textbf{66.37}~~~~~~~&\textbf{73.02}&\IEEEeqnarraystrutsize{0pt}{0pt}\\
\IEEEeqnarrayseprow[3pt]\\
\IEEEeqnarrayrulerow[1pt]\\
\end{IEEEeqnarraybox}
\vspace*{-8pt}
\end{table*} 
\begin{table*}[!t]
\centering
\renewcommand{\arraystretch}{1.3}
\setlength{\abovecaptionskip}{0.3cm}
\caption{State-of-the-art comparison results on the COCO-QA dataset.}
\label{tab_7}
\begin{IEEEeqnarraybox}[\IEEEeqnarraystrutmode\IEEEeqnarraystrutsizeadd{2pt}{2pt}]{l/c/c/c/c/c/c/c/c}
\IEEEeqnarrayrulerow[1pt]\\
\mbox{Method}~~~~~&\mbox{Faster RCNN} ~~~~~~~&\mbox{Object}~~~~~~~~&\mbox{Number}~~~~~~~~&\mbox{Color} ~~~~~~~~&\mbox{Location}~~~~~~&\mbox{WUPS0.9}~~~~~~&\mbox{WUPS0.0}~~~~~~&\mbox{Overall}\\       
\IEEEeqnarrayrulerow[0.5pt]\\    
\mbox{SAN}\mbox{\cite{b35}}~~~~~&\usym{1F5F4}~~~~~~~&64.50 ~~~~~~~~&48.60 ~~~~~~~~&57.90 ~~~~~~~~&54.00 ~~~~~~&71.60 ~~~~~~&90.90~~~~~~&61.60\\ 
\mbox{Dual-MFA}\mbox{\cite{b25}}~~~~~&\usym{2713}  ~~~~~~~&68.86  ~~~~~~~~&51.32  ~~~~~~~~&65.89  ~~~~~~~~&58.92  ~~~~~~&76.15  ~~~~~~&92.29  ~~~~~~&66.49\\            
\mbox{CVA}\mbox{\cite{b60}}~~~~~&\usym{2713}  ~~~~~~~&69.55  ~~~~~~~~&50.76  ~~~~~~~~&68.96   ~~~~~~~~&59.93  ~~~~~~&76.70  ~~~~~~&92.41  ~~~~~~&67.51\\                
\mbox{ODA}\mbox{\cite{b33}}~~~~~&\usym{2713}  ~~~~~~~&70.48  ~~~~~~~~&54.70  ~~~~~~~~&74.17   ~~~~~~~~&60.90  ~~~~~~&78.29  ~~~~~~&93.02  ~~~~~~&69.33\\  
\mbox{CoR-3}\mbox{\cite{b47}}~~~~~&\usym{2713}  ~~~~~~~&70.42  ~~~~~~~~&55.83  ~~~~~~~~&74.13   ~~~~~~~~&60.57  ~~~~~~&78.10  ~~~~~~&92.86  ~~~~~~&69.38\\ 
\mbox{ALMA}\mbox{\cite{b48}}~~~~~&\usym{1F5F4} ~~~~~~~&70.22 ~~~~~~~~&52.48 ~~~~~~~~&65.81  ~~~~~~~~&60.35  ~~~~~~&78.14  ~~~~~~&93.06 ~~~~~~&68.27 \\
\mbox{MRA-Net}\mbox{\cite{b41}}~~~~~&\usym{2713}  ~~~~~~~&71.40 ~~~~~~~~&56.42 ~~~~~~~~&74.69   ~~~~~~~~&60.62  ~~~~~~&79.03  ~~~~~~&93.21  ~~~~~~&70.27\\ 
\mbox{CAM}\mbox{\cite{b50}}~~~~~&\usym{2713}  ~~~~~~~&70.32  ~~~~~~~~&55.26   ~~~~~~~~&\textbf{77.10}  ~~~~~~~~&59.28  ~~~~~~&78.53  ~~~~~~&92.97  ~~~~~~&69.68\\
\IEEEeqnarrayrulerow[0.5pt]\\
\mbox{\textbf{LOIS (ours)}}~~~~~&\usym{1F5F4} ~~~~~~~&\textbf{72.73} ~~~~~~~~&\textbf{56.45} ~~~~~~~~&75.99  ~~~~~~~~&\textbf{62.09}  ~~~~~~&\textbf{80.19}  ~~~~~~&\textbf{94.83}~~~~~~&\textbf{71.94} \\                      
\IEEEeqnarrayrulerow[1pt]\\
\end{IEEEeqnarraybox}
\vspace*{-10pt}
\end{table*}  
 
\begin{figure*}[!t]
\centering
\includegraphics[width=7.05in]{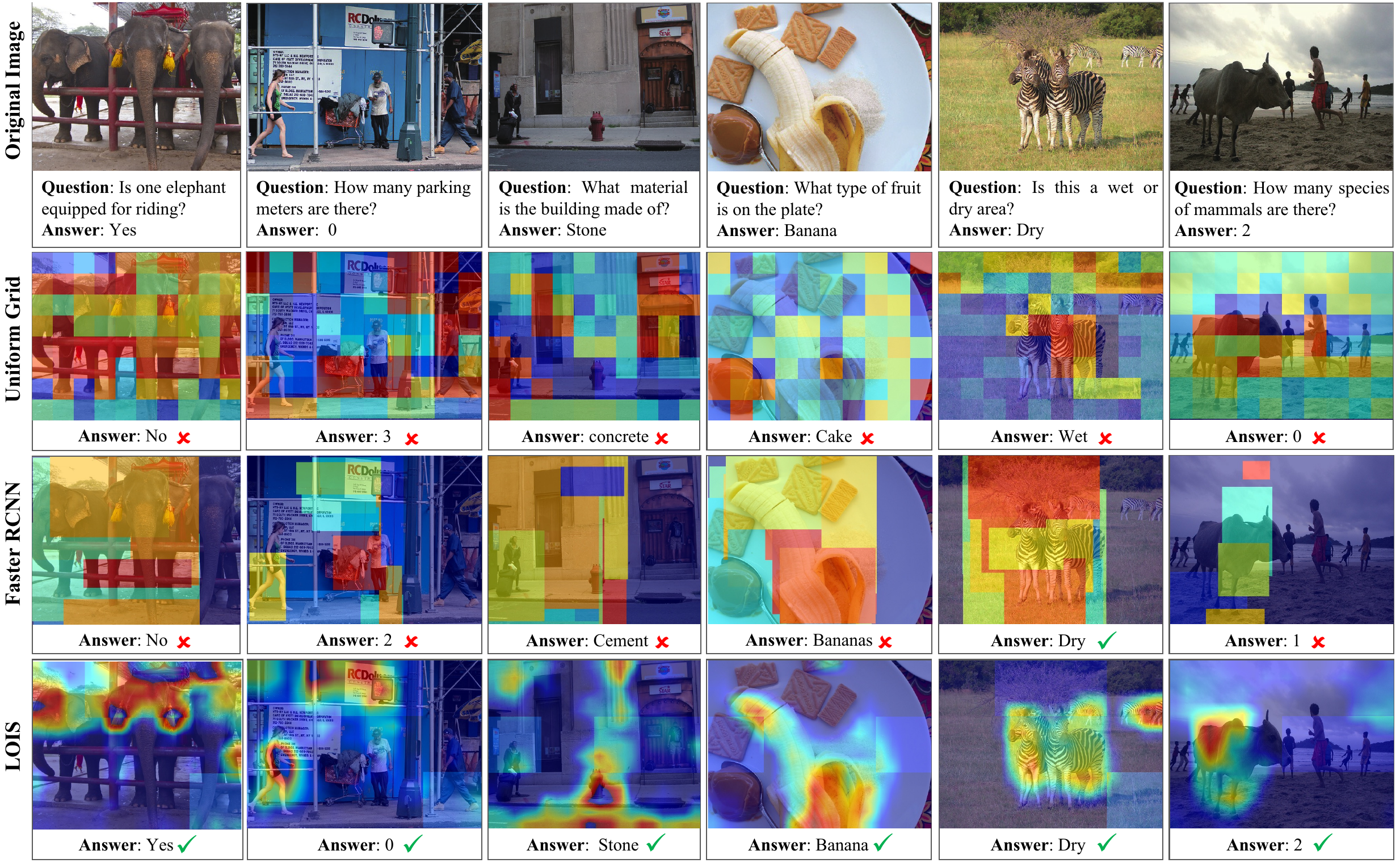}
\caption{Six examples are applied for visualization analysis. The examples contain attention maps of “Uniform Grid”, “Faster RCNN” and “LOIS”, respectively. For each instance, we provide the original image, the question-answer pair, and the answers predicted by the three methods. Notably, the samples with “\protect\usym{1F5F8}” and “\protect\usym{1F5F4}” represent correctly and incorrectly answered questions. The highlighted parts are objects with higher attention weights.}
\label{fig_11}
\end{figure*}

To evaluate whether the attention mechanisms are affected by LOIS, we compare our method with uniform grid-based and bounding box-based detection methods in Fig.~\ref{fig_11}. The models cover all question types in the VQA v2 dataset. Although the uniform grid-based attention model involves wide visual regions, it is hard to focus on one object and is easy to be interfered with by noise. Additionally, there exists a higher probability of recognizing other objects for one block in the grid. As can be observed, when answering \emph{“What type of fruit is on the plate?”}, the grid-based model cannot recognize a single complete object. Thus, failing to obtain the correct answer. The main disadvantage of the Faster RCNN-based attention model is that the object regions overlap. For instance, when answering \emph{“What material is the building made of?”}, the Faster RCNN-based model fails to separately recognize the regions related to the building, leading to the wrong answer. Different from the comparison models, our method focuses more on the interaction between the instances and background regions. Then, the second attention between the question and the image region is implemented. Taking the second instance as an example, when answering \emph{“How many parking meters are there?”}, comparison methods fail to comprehend the meaning of the question and capture the image regions related to the \emph{“parking meters”}, which is not accurate enough to infer the answer. Observably, our method predicts the correct answer \emph{“Yes”}, which can better recognize contents related to the question. Like the fifth instance, our method predicts \emph{“dry”} as a candidate answer when answering \emph{“Is this a wet or dry area?”}, which considers richer background feature information and helps the model to obtain the correct answer. For the sixth instance, taking the question of \emph{“How many species of mammals are there?”}, the Faster RCNN-based model ignores the key regions of humans. However, our LOIS model can accurately analyze and infer the correct answer. While our LOIS model predicts most of the correct regions, it still faces a serious challenge to answer the correct number. Based on this observation, these cases can help us to refine and enhance the performance of VQA.

\subsection{Comparison with State-of-the-arts}
Through a series of ablation studies, we select the proposed LOIS model that provides the best result to compare with the state-of-the-art methods on four common VQA benchmark datasets. Specifically, we evaluate the proposed LOIS on the VQA v1 dataset and compare the test-dev and test-std results to other existing works, shown in Table~\ref{tab_5}. The ALSA \cite{b39} method is conducted to learn effective multi-view features by reinforcing the two types of attention network mutually. Recent model MRA-Net \cite{b41} explore binary and trinary relations, and extract features by Faster RCNN. In contrast, our method is superior to provide visual facts. LOIS increases the overall accuracy of MRA-Net and ALSA by 2.05\% and 1.59\% on the test-dev set. Additionally, LOIS improves the performance of MRA-Net and ALSA by 1.46\% and 1.36\% on the test-std set, respectively. The above results prove that our LOIS method outperforms the compared ones by effectively improving instance object features and semantic relation reasoning.
Table~\ref{tab_6} displays the experimental results on the VQA v2 dataset. Our strategy achieves better accuracy compared to the recent state-of-the-art methods. Taking the results of the Pixel-BERT \cite{b13} method as an example, the overall accuracy of our model obtains an improvement from 71.35\% to 72.78\% on the test-dev set. For the test-std set, our overall accuracy score is 73.02\%, which improves the performance by 1.6\%. Compared with LXMERT \cite{b54}, LOIS improves 0.36\% and 0.52\% higher accuracy on test-dev and test-std, respectively. There is an improvement of 2.72\% on the test-std compared with VL-T5 \cite{b57}. Likewise, our model also outperforms UNITER \cite{b56}. In addition, we obtain a relatively higher score in the “\emph{Yes/No}”,  “\emph{Number}” and “\emph{Other}” subtasks than other comparison methods, which demonstrates that the proposed LOIS has an effective visual reasoning capability.

In Table~\ref{tab_7}, comparison results on the COCO-QA dataset are displayed. As shown in Table~\ref{tab_7}, our proposed method outperforms the other methods. Specifically, CAM \cite{b50} was a recent competitive model that explored the semantics embedded in the predicted answers by a two-stage model. Nevertheless, we improve the overall accuracy of CAM from 69.68\% to 71.94\%. Additionally, compare with other baselines, our scores of WUPS@0.9 and WUPS@0.0 are further improved.

To validate the advantages of the proposed model, we perform an extra experiment on the VQA-CP v2, which is used to overcome the question-oriented bias. LOIS has a relatively outstanding performance of 43.09\% on the “\emph{Overall}” task in Table~\ref{tab_8}. Similar improvement results can also be seen in the “\emph{Yes/No}”, “\emph{Number}”, and “\emph{Other}” categories. Significantly, we can observe that the other state-of-the-art models are more prone to question overfitting than our model.
\begin{table}[!t]
\centering
\renewcommand{\arraystretch}{1.3}
\setlength{\abovecaptionskip}{0.3cm}
\caption{Comparison results of the LOIS and several state-of-the-art methods on the VQA-CP v2 dataset.}
\label{tab_8}   
\begin{IEEEeqnarraybox}[\IEEEeqnarraystrutmode\IEEEeqnarraystrutsizeadd{2pt}{2pt}]{l/c/c/c/c/c}
\IEEEeqnarrayrulerow[1pt]\\ 
\mbox{Method}~&\mbox{Faster RCNN} ~~~&\mbox{Y/N}~~~~&\mbox{Num} ~~~~&\mbox{Other}~~~~&\mbox{Overall}\\
\IEEEeqnarrayrulerow[0.5pt]\\
\mbox{SAN}\mbox{\cite{b35}}~&\usym{1F5F4}~~~&38.35~~~~&11.14~~~~&21.74~~~~&24.96\\ 
\mbox{RAMEN}\mbox{\cite{b61}}~&\usym{2713}~~~&-    ~~~~&-       ~~~~ &-      ~~~~ &39.21\\
\mbox{MulRel}\mbox{\cite{b62}}~&\usym{1F5F4}~~~&42.85 
~~~~&13.17~~~~&45.04~~~~&39.54\\
\mbox{Bottom-up}\mbox{\cite{b26}}~&\usym{1F5F4}~~~&42.27~~~~&11.93~~~~&46.05~~~~&39.74\\
\mbox{ReGAT}\mbox{\cite{b36}} ~&\usym{2713}~~~&-    ~~~~&-       ~~~~&-   ~~~~&40.42\\
\mbox{CAM}\mbox{\cite{b50}} ~&\usym{2713} ~~~&43.29 ~~~~&12.31 ~~~~&45.41  ~~~~&39.75\\
\mbox{MRA-Net}\mbox{\cite{b41}} ~&\usym{2713} ~~~&44.53 ~~~~&13.05 ~~~~&45.83  ~~~~&40.45\\
\mbox{MLVQA}\mbox{\cite{b59}} ~&\usym{2713} ~~~&40.84 ~~~~&13.24 ~~~~&\textbf{48.83}  ~~~~&41.08\\
\IEEEeqnarrayrulerow[0.5pt]\\
\textbf{\mbox{LOIS (ours)}} ~~&\usym{1F5F4}~~~&\textbf{51.82} ~~~~&\textbf{14.24} ~~~~&41.39 ~~~~&\textbf{43.09}\\
\IEEEeqnarrayrulerow[1pt]\\
\end{IEEEeqnarraybox}
\vspace*{-15pt}
\end{table} 
\section{Conclusion}
In this work, we propose LOIS, a more fine-grained semantic detector framework for visual question answering. By separating foreground objects and background forms, we emphasize the non-negligible importance of edge feature distribution. Moreover, to correctly infer the complex semantic correlations between the salient image regions and the question, we explore visual bottleneck at the multi-view level by reconciling the intra-modal and inter-modality relation attention modules. To effectively evaluate the performance of the presented model, we compare it with the state-of-the-art strategies by conducting extensive experiments on four benchmark VQA datasets. The experimental results demonstrate that the proposed method is better at capturing the high-level interactions between the vision and language domains. The findings of this research raise two points for further consideration. One is that the designed LOIS framework will be applied to more VQA scenarios and tasks in the future, and the other is to further explore the different semantic correlations between the images and questions.
 

\vspace{-14 pt}

\end{document}